\documentclass[sn-mathphys,Numbered]{sn-jnl}

\usepackage{graphicx}%
\usepackage{multirow}%
\usepackage{amsmath,amssymb,amsfonts}%
\usepackage{amsthm}%
\usepackage{mathrsfs}%
\usepackage[title]{appendix}%
\usepackage{xcolor}%
\usepackage{textcomp}%
\usepackage{manyfoot}%
\usepackage{booktabs}%
\usepackage{algorithm}%
\usepackage{algorithmicx}%
\usepackage{algpseudocode}%
\usepackage{listings}%
\usepackage{caption} 
\usepackage{multirow}
\usepackage{float}
\usepackage{url}
\usepackage{booktabs} 
\usepackage{siunitx} 

\theoremstyle{thmstyleone}%
\theoremstyle{thmstyletwo}%

\theoremstyle{thmstylethree}%

\begin{document}

\title[Article Title]{Leveraging Natural Language Processing For Public Health Screening On YouTube: A COVID-19 Case Study}


\author*[1]{\fnm{Ahrar Bin Aslam} \sur{}}\email{18tl04@students.muet.edu.pk}

\author*[2]{\fnm{Zafi Sherhan Syed} \sur{}}\email{zafisherhan.shah@faculty.muet.edu.pk}

\author[3]{\fnm{Muhammad Faiz Khan} \sur{}}\email{18tl116@students.muet.edu.pk}
\equalcont{These authors contributed equally to this work.}

\author[4]{\fnm{Asghar Baloch} \sur{}}\email{18tl76@students.muet.edu.pk}
\equalcont{These authors contributed equally to this work.}

\author[5]{\fnm{Muhammad Shehram Shah Syed} \sur{}}\email{shehram.shah@faculty.muet.edu.pk}
\equalcont{These authors contributed equally to this work.}

\affil*[1]{\orgdiv{Department of Telecommunication}, \orgname{Mehran University of Engineering and Technology}, \orgaddress{\state{Jamshoro}, \country{Pakistan}}}

\affil*[2]{\orgdiv{Department of Telecommunication}, \orgname{Mehran University of Engineering and Technology}, \orgaddress{\state{Jamshoro}, \country{Pakistan}}}

\affil[3]{\orgdiv{Department of Telecommunication}, \orgname{Mehran University of Engineering and Technology}, \orgaddress{\state{Jamshoro}, \country{Pakistan}}}

\affil[4]{\orgdiv{Department of Telecommunication}, \orgname{Mehran University of Engineering and Technology}, \orgaddress{\state{Jamshoro}, \country{Pakistan}}}

\affil[5]{\orgdiv{Department of Software Engineering}, \orgname{Mehran University of Engineering and Technology}, \orgaddress{\state{Jamshoro}, \country{Pakistan}}}


\abstract{\textbf{Background:} Social media platforms have become a viable source of medical information, with patients and healthcare professionals using them to share health-related information and track diseases. Similarly, YouTube, the largest video-sharing platform in the world contains vlogs where individuals talk about their illnesses. The aim of our study was to investigate the use of Natural Language Processing (NLP) to identify the spoken content of YouTube vlogs related to the diagnosis of Coronavirus disease of 2019 (COVID-19) for public health screening.

\textbf{Methods:} COVID-19 videos on YouTube were searched using relevant keywords. A total of 1000 videos being spoken in English were downloaded out of which 791 were classified as vlogs, 192 were non-vlogs, and 17 were deleted by the channel. The videos were converted into a textual format using Microsoft Streams. The textual data was preprocessed using basic and advanced preprocessing methods. A lexicon of 200 words was created which contained words related to COVID-19. The data was analyzed using topic modeling, word clouds, and lexicon matching. 

\textbf{Results:} The word cloud results revealed discussions about COVID-19 symptoms like ``fever'', along with generic terms such as ``mask'' and ``isolation''. Lexical analysis demonstrated that in 96.46\% of videos, patients discussed generic terms, and in 95.45\% of videos, people talked about COVID-19 symptoms. LDA Topic Modeling results also generated topics that successfully captured key themes and content related to our investigation of COVID-19 diagnoses in YouTube vlogs.

\textbf{Conclusion:} By leveraging NLP techniques on YouTube vlogs public health practitioners can enhance their ability to mitigate the effects of pandemics and effectively respond to public health challenges. 
}




\maketitle

\section{INTRODUCTION}\label{sec1}

Coronavirus is an infectious disease caused by the SARS-CoV-2 virus~\cite{Coronavirusdisease(COVID-19)} which was first identified in December 2019 in Wuhan, China. Since then, it started spreading all around the world and was declared a pandemic by the World Health Organization (WHO) on 11th March 2020~\cite{Coronavirusdisease(COVID-19)1}. This disease was termed COVID-19 by the WHO~\cite{Coronavirusdisease(COVID-19)2}. COVID-19 transmits when people breathe in air contaminated by droplets and small airborne particles. Individuals afflicted with this ailment experience a diverse array of symptoms~\cite{COVID-19Symptoms} such as fever, cough, shortness of breath, headache, sore throat, muscle ache, congestion, loss of sense of taste and smell, fatigue, nausea, and diarrhea.

The preventive measures for COVID-19 are getting vaccinated, wearing a mask, maintaining at least a 1-meter distance from others, and keeping good hygiene~\cite{AdviceforPublicforCOVID-19}. As of 24 May 2023, there have been 766,895,075 confirmed cases since the start of the pandemic, out of which 6,935,889 cases have resulted in deaths of the patients~\cite{COVID-19Statistics}. The COVID-19 pandemic has presented numerous challenges for public health worldwide, with a significant burden on healthcare systems. The demand for basic healthcare materials such as ICU beds, ventilators, and personal protective equipment (PPE) has outpaced the supply in many hospitals and medical facilities. In addition, the pandemic has resulted in a shortage of healthcare personnel, including doctors and nurses, as they work tirelessly to combat the virus~\cite{el2020assessing}. In global response to the pandemic, ensuring enough availability and adequate distribution of COVID-19 vaccines have emerged as a critical concern. Notably, developing countries have faced serious challenges in accessing vaccines, which has resulted in big differences in immunization coverage among different populations. This has resulted in healthcare professionals being exposed to the virus and dying~\cite{singh2021prioritizing}.

This pandemic also affected the mental health of the public, Vindegaard et al.~\cite{vindegaard2020covid} explored the influence of COVID-19 on the mental well-being of both, individuals who had been infected with the virus and the public, which also includes healthcare professionals and people who had preexisting mental health issues. It was found that the infected patients had an elevated level of post-traumatic stress symptoms (PTSS) and an increasingly higher level of depressive symptoms. Whereas the public including healthcare workers and people with preexisting mental health issues also observed an increased level of anxiety and depressive symptoms. Qiu et al.~\cite{qiu2020nationwide} revealed the psychological distress caused by COVID-19 on individuals in China. The study aimed to investigate the impact of the pandemic on the mental health of the Chinese population. In our prior study~\cite{10007300}, we examined the influence of COVID-19 on emotions and sentiments. Our findings indicate that there was a rise in negative emotions at the beginning of the pandemic when it was officially declared. We did observe that it took several months for emotions to return to their pre-pandemic levels. This suggests that the psychological impact of the pandemic was long-lasting and had a considerable effect on the emotional health of the individuals.

Lately, there has been a rise in the utilization of social media for obtaining crucial information about public health given that the public uses social media platforms every day to share their health-related information~\cite{kass2013social}. Due to COVID-19 being a pandemic in this digital age, there is a large amount of data available on different social media platforms where users are sharing their experiences and journey related to the pandemic. Social media platforms have proved to be pivotal in spreading information regarding this pandemic~\cite{gonzalez2020social}. In the recent past, one finds several studies which have proposed leveraging social media posts to understand and analyze disease-related information for public health purposes. For example, posts on Reddit were used by Park et al.~\cite{park2020insights} to discern mental health trends in patients suffering from rheumatoid arthritis. Similarly, Lian et al.~\cite{lian2022} identified adverse effects of the COVID-19 vaccine from Twitter data. In addition, researchers have also used YouTube to analyze the content quality of different diseases~\cite{onder2021youtube, godskesen2021youtube, szmuda2020youtube}. 

Since the pandemic first began, people have utilized social media platforms to share their experiences, symptoms, and recovery process related to this virus. A large amount of data available on people’s experiences presents us with an opportunity to analyze the methods of COVID-19 treatments and care.
As YouTube is the largest video-sharing platform in the world, having over 2 billion monthly logged-in users~\cite{YoutubeStatistics}, it offers significant potential for research in the field of public health. We argue and show that it can be utilized for public health screening of COVID-19. 

\subsection{Motivation}\label{sec2}

The COVID-19 pandemic has attracted a lot of public interest, with people sharing their experiences of symptoms and recovery from this virus on social media platforms such as YouTube. This offers an opportunity to contribute to present and future public health screening efforts aimed at fighting pandemics. Leveraging COVID-19 related data on YouTube, an automatic system could be created using Automatic Speech Recognition (ASR) tools to analyze speech and match symptoms with those in this data to determine if an individual has COVID-19. This approach could potentially detect COVID-19 at an early stage, enabling prompt public health response to stop the spread of this virus. Overall, content analysis of COVID-19 related vlogs on YouTube and ASR tools could prove to be a productive means of enhancing public health surveillance efforts during the COVID-19 pandemic.

\subsection{Research Questions}\label{sec3}

Around the world, anxiety and depression have risen substantially because of the COVID-19 pandemic~\cite{COVID-19IncreaseinAnxiety}, and initially, even the health care workers (HCW) faced a shortage of reliable information about this virus~\cite{bhagavathula2020knowledge}. As more information became available, knowledge about COVID-19 symptoms and their impact on public health has been gained through patient history. Our research seeks to investigate whether natural language processing (NLP) methods can be used to examine transcripts of COVID-19 related YouTube vlogs, which may contain similar patient histories. Specifically, we seek to address the following questions. \newline
RQ1: Can we use NLP to facilitate the identification of COVID-19 symptoms from such vlogs?
\newline
RQ2: Which traits can NLP identify from YouTube vlogs that can be utilized for public health screening of viral
diseases?

\subsection{YouTube as a Source of Public Health Surveillance}
Our research aspires to contribute to the present efforts to combat the COVID-19 pandemic by analyzing YouTube vlogs of people’s personal experiences of being infected with this virus. To the best of our knowledge, this is the first study that has explored the potential of analyzing COVID-19 related YouTube vlogs for identifying patterns and themes related to symptoms experienced by individuals with COVID-19. This represents an important contribution to the field of public health surveillance, as YouTube is a widely used platform for people to share their personal experiences and knowledge of COVID-19, and the analysis of such vlogs could provide valuable insights for public health officials and practitioners.

The rest of the paper is organized as follows: Section 2 presents a brief overview of the related work. Section 3 provides a detailed description of the dataset used in this research and the preprocessing steps implemented to clean the data. The NLP methods used in extracting key information regarding COVID-19 are described in Section 4. The results obtained from these experiments are presented in Section 5. The discussion, key takeaways and insights, limitations, implications of this research, and directions for future work are presented in Section 6. 

\section{RELATED WORK}
\label{Relatedwork}
Natural Language Processing (NLP) is a branch of artificial intelligence that deals with the interactions between computers and human language. It is concerned with the understanding and interpretation of human language, and how to program computers to analyze, understand, and derive meaning from what humans are speaking. Recent developments in NLP technologies are opening new possibilities for health research and decision-making based on the analysis of large volumes of text quickly. 

Utilizing the data available in scientific literature, health records, technical reports, social media, surveys, etc. can help to facilitate the public health system and improve the current surveillance systems by identifying the diseases faster, whether or not there are any risk factors associated with the disease, to get the before-hand knowledge in-case the disease could turn out be an outbreak or a pandemic, disease prevention strategies and getting accurate answers regarding various health issues and queries. NLP is becoming recognized as a crucial tool that public health officials can use to help lessen the burden of health inequality and unfairness in the population~\cite{baclic2020artificial}. A study was conducted by Al-Garadi et al.~\cite{al2022role} to identify the applications and usefulness of natural language processing in present and forthcoming pandemics. By applying NLP to many pandemic specific problems their research highlighted that its methods can serve as a useful tool to improve pandemic preparedness and response.

In the recent past, one finds several studies in which scientists have utilized NLP methods to analyze social media data for public health~\cite{conway2019recent}. Since the outbreak of this virus, the world went into lockdown and people were not able to meet their friends, colleagues, and relatives. As a result, people started to feel alone and uncertain regarding the outcome of the pandemic~\cite{polvsek2020huremovic}. Having ample time in their homes and with the hysteria surrounding this disease, e.g.\textit{ How deadly is it? What effects would it have on our lives?} People started sharing their experiences related to COVID-19 in the form of videos, posts, and tweets on different social media platforms. Because of that, we have a large amount of data on social media of COVID-19 available. 

It has been shown that NLP can be used to investigate various aspects of COVID-19 through data acquired from different social media platforms ~\cite{liu2021monitoring,low2020natural,wicke2021covid,patel2021analysis,bose2021comparative}. Our study draws inspiration from the following papers, as we have also used NLP in examining COVID-19 related content, however, our study focused on videos available on YouTube distinguishing it from others that examined COVID-19 related content on other social media platforms.

A study by Liu et al.~\cite{liu2021monitoring} aimed to assess the effectiveness of Reddit as a platform for COVID-19 pandemic surveillance, with an emphasis on North Carolina. They used natural language processing techniques such as Bidirectional Encoder Representations from Transformers (BERT)-based sentence clustering, named-entity recognition, cosine similarity measures, and Latent Dirichlet allocation (LDA) topic modeling to analyze Reddit posts from March to August 2020. The dataset was broken down into two periods, and LDA topic modeling was used to identify 5 topics for each period. To preprocess the data URL removal, tokenization, punctuation, stop-word removal, part-of-speech tagging, and lemmatization were performed. The study shed light on people's concerns and behaviors in North Carolina during the COVID-19 pandemic.

In~\cite{low2020natural}, Low et al. employed natural language processing on Reddit posts from 2018 to 2020 to investigate changes in non-mental health communities and major mental health support groups during the initial stages of the pandemic. They used supervised machine learning to classify posts into appropriate support groups and unsupervised techniques like topic modeling and clustering to identify issues throughout Reddit. Posts from fifteen subreddits were analyzed, and data from both pre-pandemic and mid-pandemic periods were used for classification, clustering, and topic modeling. Pre-processing involved removing stop words and tokenization using NLTK (Natural Language Toolkit) Tweet Tokenizer, while LDA topic modeling was used to analyze the topics in the textual dataset. The study found an increase in the distribution of the Health Anxiety topic across mid-pandemic posts, along with changes in other topics between pre- and mid-pandemic periods.

An experimental analysis was conducted by Philipp Wicke and Marianna M. Bolognesi~\cite{wicke2021covid} on how the conversation on Twitter surrounding COVID-19 developed during the initial wave of the pandemic. They utilized topic modeling on a large corpus of tweets created between 20 March and 1 July 2020 to show the evolution of subjects associated with the spread of the pandemic over time. The language used in tweets also shifted from positive to negative valence in response to reopening, while the average subjectivity of tweets increased linearly. Additionally, they examined how the widely used metaphor for war was altered by actual riots and street clashes. To create a sub-corpus of COVID-19 related tweets, they used the resource of Lamsal~\cite{Lamsal}, which contained 3-4 million tweets per day in English from Twitter based on more than 90 keywords related to COVID-19.

To investigate how the COVID-19 pandemic affected people's physical and mental health, Patel et al.~\cite{patel2021analysis} collected data from three online health forums with a wide audience and regular activity. Using lemmatization, 28 keywords were categorized into three groups: physical symptoms, intensive care terms, and mental health symptoms. Posts were searched for these keywords and their prevalence throughout the pandemic was analyzed. Data was standardized for unicode, whitespace, date, time, and location of posters. The study aimed to determine changes in the prevalence of worries about physical and mental health symptoms and intensive care terms throughout the pandemic. Analysis was done using bespoke Python software.

The study conducted by Bose et al.~\cite{bose2021comparative} aimed to examine the effects of COVID-19 on various aspects of society, such as social and economic factors, by utilizing natural language processing techniques on research literature. They collected 10,000 abstracts from PubMed on four diseases, namely COVID-19, Influenza, Middle East Respiratory Syndrome (MERS), and Ebola, and characterized them into ten topics relating to several aspects of the diseases. The study compared COVID and non-COVID abstracts and discovered that PubMed abstracts contain a significant amount of extraneous information that needs to be removed. After cleaning and processing the data, word clouds were generated for each disease, and common words were compared. Latent semantic analysis was also applied to identify the co-occurrence of MERS and COVID-19 in scientific studies. The study highlights the need for achieving herd immunity and problems relating to drugs.

\subsection{Limitations of Existing Approaches}
Several limitations were found in the studies discussed in the above section. In the research of Lui et al.~\cite{liu2021monitoring}, the data collection was limited to a specific period between March 3, 2020, and August 31, 2020, and only included English language postings from six North Carolina communities, which may not be representative of all areas. Additionally, the authors were unable to ensure that the posters were residents of the region at the time of posting. In Low et al. study~\cite{low2020natural}, the lack of formally documented clinical diagnosis was a limitation, as well as the inability to establish a relationship between language changes and events during the pandemic. The research of Patel et al.~\cite{patel2021analysis} was limited in its ability to determine whether posters had COVID-19 or preexisting mental health issues, and it was concluded that the usefulness of online health forums is up to individual discretion. In the study of Bose et al.~\cite{bose2021comparative}, the search was limited to the first 10,000 results for each search topic, which may not have been representative of all relevant literature. Additionally, the similarity between disease terms and the lack of research on COVID-19 vaccines were identified as challenges. Lastly, the topic modeling results did not highlight topics related to the impact of COVID-19 on mental, social, and economic issues.

\section{DATASET}
\subsection{Data Source and Collection}
The data collection process for this study spanned from January 2022 to September 2022 and involved three team members. To gather relevant COVID-related videos from YouTube, a set of keywords was used, including ``I got COVID'', ``My COVID experience'', ``My coronavirus experience'', ``My COVID story'', ``I survived COVID'', ``COVID got me'', ``My day-by-day COVID symptoms'', ``My Omicron experience'', ``My Delta variant experience'', and ``My COVID recovery.'' The search included videos posted since the onset of the COVID-19 pandemic in January 2020, up until September 2022. No specific criteria were established for the length of the videos, so both short (less than 5 minutes) and long (more than 30 minutes or 1 hour) videos were included. An online YouTube video downloader~\cite{ensavefromnet} was used to download the videos in 360p and 720p format. 

Notably, this data collection process was done carefully and diligently to ensure that the videos had relevant information to the topic. This approach allowed the team to compile videos that provide key insights into the personal experiences of individuals with COVID-19.

\subsection{Dataset Size and Composition}
Based on the method of data collection for this study, 1000 videos were downloaded from YouTube. These videos were then classified into three distinct categories: vlogs, non-vlogs, and deleted videos. Out of these 1000 videos, 791 belonged to the vlog category, 192 were identified as non-vlogs, and 17 videos were deleted by their respective channels after they were downloaded. The categorization of the downloaded videos is presented in a tabular format in Table \ref{table1}.

To ensure the quality and relevance of the data for analysis, only vlogs were included for further processing as vlogs are most likely to have a person’s relevant experience with the pandemic and would supply more valuable insights as compared to an interview which would also contain dialogues from the interviewer. By selecting vlogs, a rich source of information could be used to derive meaningful insights into the experiences of individuals with COVID-19. 

\begin{table}[htp]
\caption{Categorization of downloaded COVID-19 videos}
\captionsetup{justification=centering} 
\begin{tabular}{cc}
\toprule
\textbf{Categories}       & \textbf{Number of videos} \\
\midrule
Vlogs                     & 791                       \\
Non-Vlogs                 & 192                       \\
Videos which were deleted & 17                        \\
\bottomrule
\end{tabular}
\label{table1}
\end{table}

\subsection{Automatic Speech Recognition}
The videos were converted into a textual format using the Automatic Speech Recognition tool, Microsoft Streams. In our recent study~\cite{syed2021tackling}, we found that Microsoft Streams is an effective tool to generate transcripts from video files. This tool converts videos in .MP4 format into .VTT files, which are text files saved in the Web Video Text Tracks format. VTT files include additional information about a web video, including chapters, subtitles, captions, and metadata. They are commonly used for video subtitling~\cite{VTT}. The downloaded transcript files from Microsoft Stream contained the keyword \emph{Autogenerated Caption} at the end of their filename, which was then removed from all the transcripts in the dataset.
\subsection{Data Annotation}
To label the data, data annotation was performed using Microsoft Excel. Each video was described, including what the speaker was saying, the name of the channel, the link of the video, and classified into three categories: vlogs, non-vlogs, and videos that were deleted by the channel after being downloaded. The criteria for a video to be considered a vlog was that it should feature a person sharing their COVID-19 experience in their home, office, or park without any irrelevant information from a second speaker. Whereas interviews and podcasts where news anchors and Youtubers were asking questions from individuals about their experience with this virus were considered as non-vlogs. 
Throughout the data collection process, 17 videos were downloaded during our data collection process, however, they were deleted afterward from YouTube by the user, mainly because they felt shy about sharing their personal information. 

\subsection{Data Preprocessing}
The transcript files before preprocessing contained irrelevant information that did not contribute to our analysis. These included parameters such as note duration, note recognizability, note language, and note confidence. It also displayed the time span of the sentences being spoken. As we only needed the text for further processing, we excluded such irrelevant parameters. We used two types of preprocessing settings, Basic Preprocessing and Advanced Preprocessing. The two methods have been described in the following sections. 
\subsubsection{Basic Preprocessing}
\begin{figure}[htp]
  \centering 
  \includegraphics[width=1.0\textwidth]{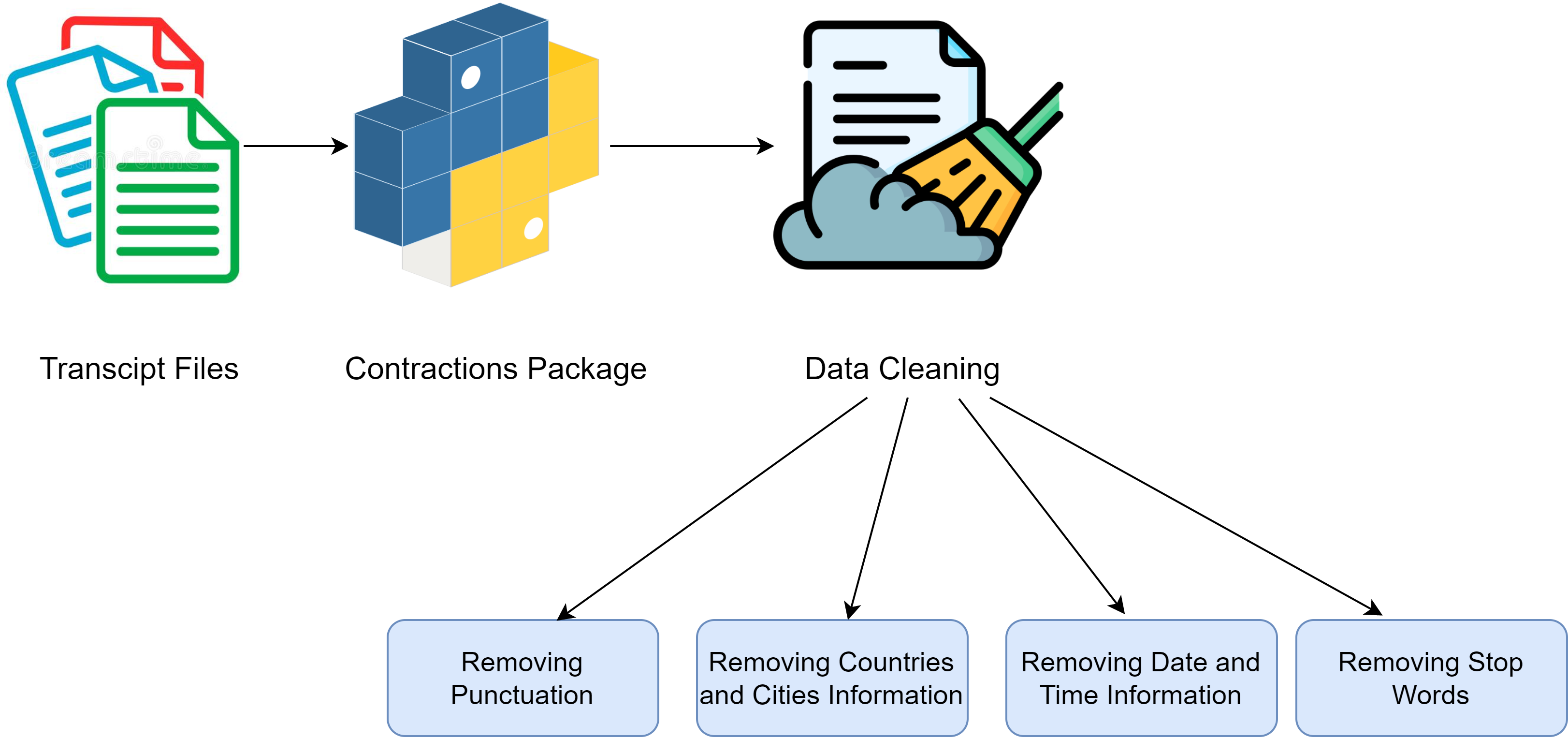} 
  \captionsetup{justification=centering} 
  \caption{Basic Preprocessing}
  \label{Basic}
\end{figure}

In this setting, we started by converting all characters in the transcripts to lowercase. After that contractions package~\cite{pythoncontractions} was used to expand words that had contractions within them. This step was used before the removal of all types of punctuation since the latter would have removed punctuations that are otherwise part of contractions.  
After these three steps, the next step was to remove time and date information from the transcripts. We noticed through data analysis that several YouTube vloggers had mentioned such information in their vlogs. Since this information was not useful for the task at hand, we removed it. Similarly, we also removed information about cities and countries from the transcripts since that too was not deemed to be useful. 
In addition to these steps, we also remove stop words from the transcripts. This is a common preprocessing step for topic modeling where the aim is to reduce the vocabulary size of the corpus with the added benefit of increasing the specificity of the words to the task at hand within the corpus. The flow diagram of basic preprocessing is represented in Figure~\ref{Basic}.
\renewcommand{\figurename}{Figure}
\newline
\subsubsection{Advanced Preprocessing}
\begin{figure}[htp]
  \centering 
  \includegraphics[width=1.0\textwidth]{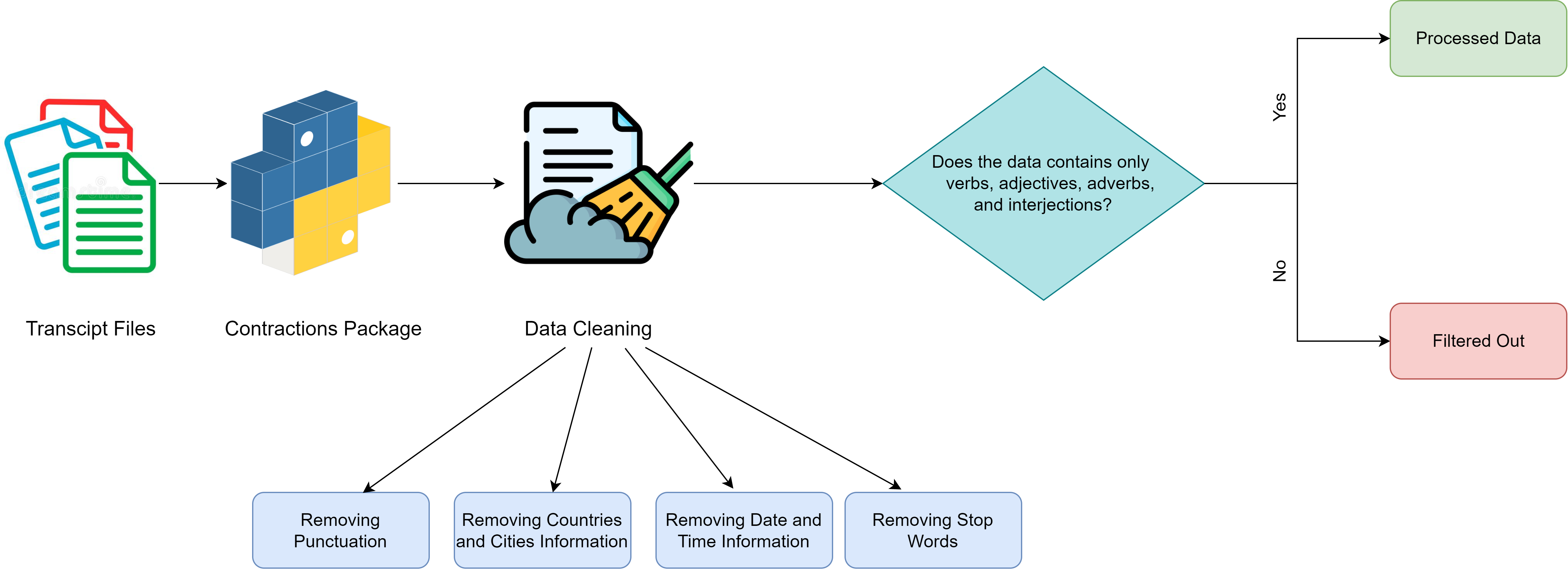} 
  \captionsetup{justification=centering} 
  \caption{Advanced Preprocessing}
  \label{Advanced}
\end{figure}

In addition to the steps involved in basic preprocessing, we only retained tokens that belong to one of the four Parts-of-Speech tags i.e., "ADJ" (adjective), "VERB" (verb), "ADV" (adverb), and "INTJ" (interjection) as presented in Figure 2. We surmised that adjectives, adverbs, interjections, and verbs may carry the most semantic information when a text document is processed for public health screening. 
As an example, consider the following fictitious transcript: \emph{"covid was quite bad for me since I was first unsure about the virus, and everyone was going into lockdown. I got sick and I thought I was not going to make it. I took painkillers and other medicine and eventually recovered thank goodness"}. Here, the adjectives "sick" and "unsure" provide information about the vloggers' health and emotional state, respectively, and the verbs "got" and, "thought" provides information about actions performed, and the adverb "eventually" provides information about a time. Based on this understanding, one can surmise that topic modeling is more likely to correctly identify the themes that are relevant to public health screening. We aimed to retain the most semantically rich information about the text. The flow diagram for advanced preprocessing is illustrated in Figure~\ref{Advanced}.

\begin{figure}[H]
  \centering
  \includegraphics[width=1.0\textwidth]{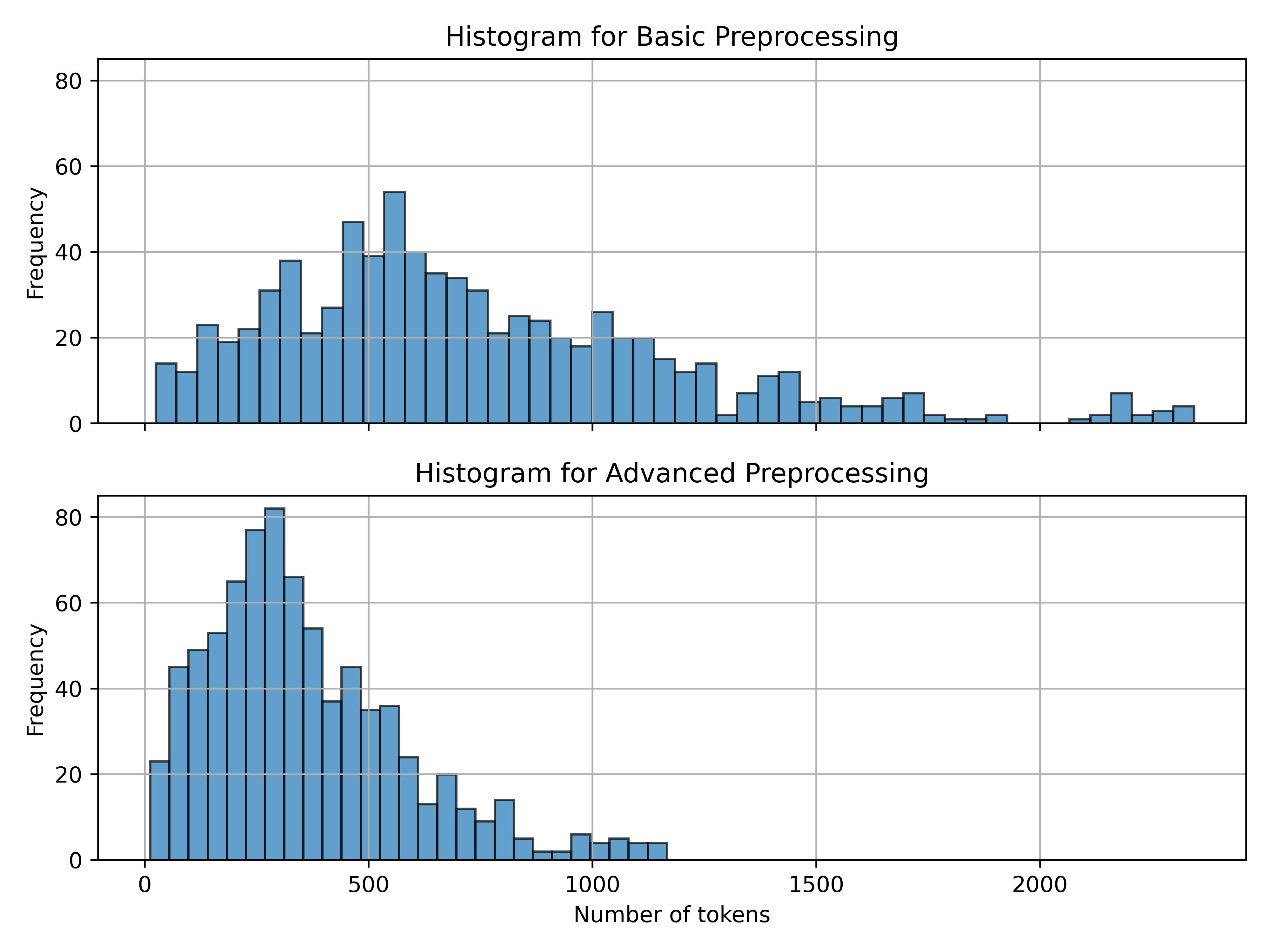}
  \caption{Histogram of Preprocessing Methods}
  \label{Histogram}
\end{figure}

Our secondary motivation for exploring the Advanced Preprocessing setting was that it will retain only relevant words thereby reducing the size of the corpus. As illustrated in Figure~\ref{Histogram}, the number of tokens in the Advanced Preprocessing method is significantly smaller than in the Basic Preprocessing setting. 
\renewcommand{\figurename}{Figure}
\section{METHODS}

\subsection{Overall Approach}
\begin{figure}[H]
\centering 
\includegraphics[width=1.0\textwidth]{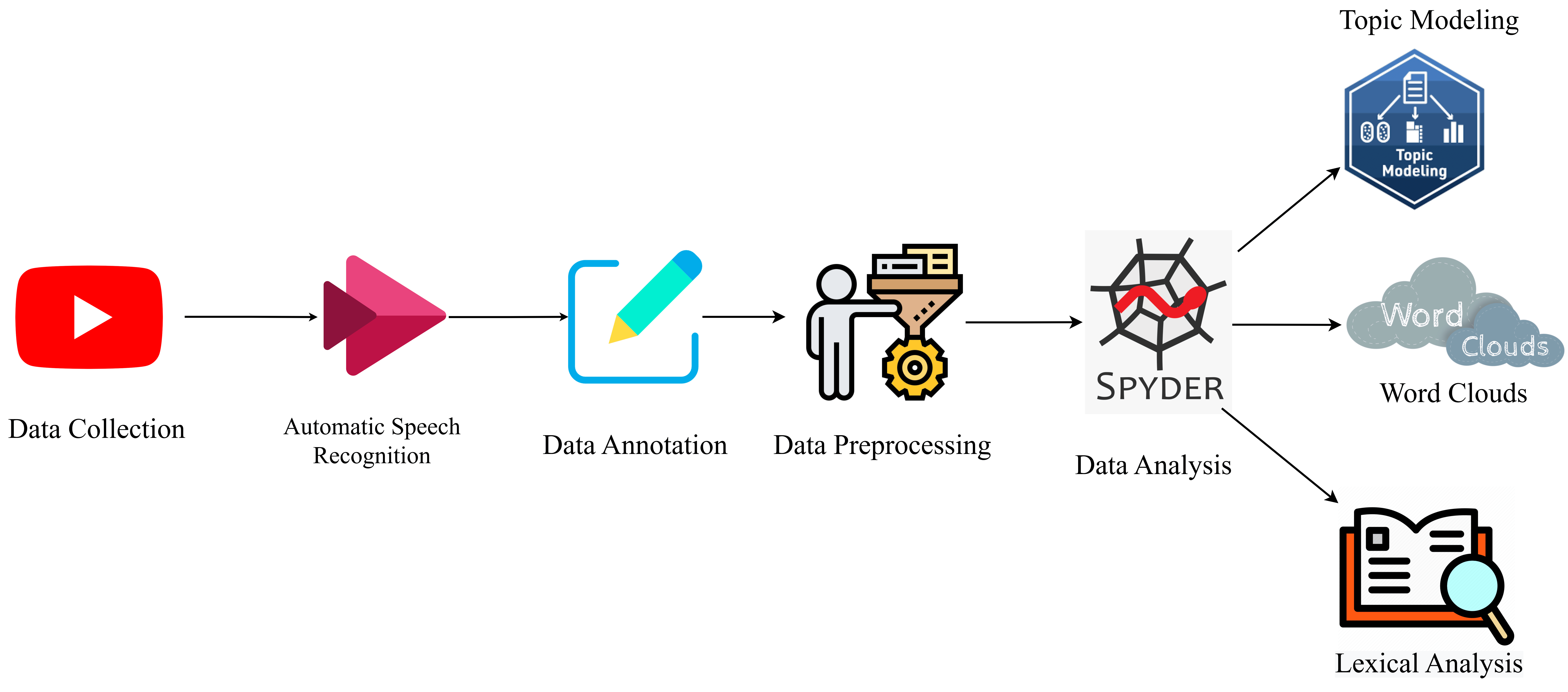} 
\captionsetup{justification=centering} 
\caption{Process Flow Diagram of the Methodology}
\label{Methodology}
\end{figure}

Now we shall define the methodology for various aspects of our experiments that includes dataset collection, curation, and preprocessing. We shall also talk about how we analyzed the data using various natural language processing methods. The process flow diagram of the methodology is presented in Figure~\ref{Methodology}.
\renewcommand{\figurename}{Figure}

\subsection{COVID-19 Lexicon Matching}
Lexicon is a specialized vocabulary for a particular topic and is essential for retrieving and interpreting text-based clinical information~\cite{liu2012towards}. It contains semantic and grammatical data about specific words or word groups and is part of the NLP system~\cite{guthrie1996role}. Since the start of the pandemic, there have been several terms associated with COVID-19, for instance, isolation, quarantine, immunity, etc. In the data that we have collected, people, healthcare professionals, and government officials have used such terms while sharing their experiences with this disease. To analyze what sort of words most people speak related to this pandemic, we have created a lexicon of COVID-19 consisting of 200 words. The lexicon includes words that have been commonly used during the pandemic and is intended to be a valuable resource for future research on this topic. These terms have been grouped into 6 categories as shown in Table~\ref{table2}. We matched our lexicon with our processed transcripts to inspect what percentage of words in the transcripts belong to how many categories. We shall provide this lexicon for future research on this topic. 
\begin{table}[htp]
\caption{Categorization of Lexicon Terms}
\captionsetup{justification=centering} 
\begin{tabular}{cc}
\toprule
\textbf{Categories} & \textbf{Number of words} \\
\midrule
Generic Terms       & 113                      \\
Symptoms            & 33                       \\
Variants and Names  & 15                       \\
Medical Equipment's & 14                       \\
Medicines           & 11                       \\
Vaccines            & 8                        \\
\bottomrule
\end{tabular}
\label{table2}
\end{table}

The inclusion criteria for the lexical terms were that they had to be in English, relevant to COVID-19, and released by well-known health or governmental websites. Table~\ref{table3} displays the listing of words and their links.

\begin{table}[htp]
\caption{Lexicon terms and their links}
\captionsetup{justification=centering} 
\begin{tabular}{cc}
\toprule
\textbf{Sources}                   & \textbf{Number of Words} \\
\midrule
WebMD                              & 92                       \\
relief.unboundmedicine             & 28                       \\
TMC Houston                        & 19                       \\
UVA Health                         & 11                       \\
Mayo Clinic                        & 10                       \\
covid19.trackvaccines.org/Pakistan & 7                        \\
Yale Medicine                      & 6                        \\
CDC                                & 6                        \\
CDHN Ireland                       & 3                        \\
Healthdirect.gov.au                & 3                        \\
ec.europa.eu                       & 3                        \\
nps.org.au                         & 2                        \\
centerforhealthsecurity.org        & 2                        \\
apps.who.int                       & 2                        \\
Ominia-Health                      & 2                        \\
rochester.edu                      & 1                        \\
BBC                                & 1                        \\
FDA                                & 1                        \\
metaqil                            & 1                        \\
\bottomrule
\end{tabular}
\label{table3}
\end{table}
\subsection{Word Cloud Generation}
Word clouds are a visual representation of textual data, it contains a cloud of words from processed text files, and the more frequently a word occurs the bigger and bolder it would show up. These are excellent methods for extracting the most important portions of textual material, including blog posts and databases, and are also referred to as tag clouds or text clouds. 

To compute the word clouds, we used Python programming language. A package of word clouds was installed and various NLP libraries i.e., Spacy, NLTK, Scikit-learn were imported. To ensure the accuracy of our analysis, we removed stop words, which are commonly used in the English language and do not contribute to our analysis, using the list \emph{spacy.lang.en.stop\_words} from the Spacy library The text was then cleaned by converting it into a list by using the \emph{Text.tolist()} function, and emojis were removed as they do not provide significant information in the text-based analysis. We imported the Natural Language Toolkit (NLTK) library suite which is a collection of libraries and applications for statistical language processing~\cite{loper2002nltk}. Being one of the most prominent NLP libraries, it includes tools that allow computers to understand human language and respond appropriately when it is used. Further, we used the \emph{WordNetLemmatizer().lemmatize} method from the NLTK library to lemmatize the text. For computing the word clouds, we employed a statistical technique called Term Frequency Inverse Document Frequency (TFIDF) which assesses how pertinent a word is to a document within a collection of documents~\cite{ramos2003using}. We created a function that took input and tokenized it using the output of the lemmatized method. We then used \emph{TfidfVectorizer}, a class in the scikit-learn library to convert a collection of raw documents into a matrix of TF-IDF features. This class included the \emph{ngram\_range} which specified the order of the word cloud that needed to be computed for instance, (2,2), (3,3), (4,4) referred to as bi-gram, tri-gram, and tetra-gram respectively. To fit the vectorizer to the input text and transform the text into a matrix of TF-IDF features, we used the \emph{fit\_transform} method of the vectorizer object. We calculated the list of TF-IDF weights containing the word and their respective weights, which was created by iterating over the vocabulary of the vectorizer object and adding up the TF-IDF weights of each word across all the documents. This list gives an idea of the importance of each word in the input text, with higher weights indicating more importance. By using this method, we computed n-gram word clouds. The results of these computations are discussed in the experiments and results section. 
\subsection{Topic Modeling using LDA}
Topic Modeling is an unsupervised machine learning technique that can scan a group of documents, find word and phrase similarities among them, and then automatically group words and phrases into topics that best describe the group of documents. There are various methods of performing topic modeling and for our project, we have employed Latent Dirichlet Allocation (LDA)~\cite{blei2003latent}. It is used for text analysis to discover hidden topics by assuming potential topics from words in the documents. It uses Dirichlet distributions and a generative probabilistic model to map documents to a list of topics and assigns each word in a document to a separate topic. The algorithm computes a predefined number of topics, assigns probabilities to each word for each topic, and identifies the top words with the highest probabilities to represent those topics. Various research has been conducted in the past that has implemented LDA topic modeling for extracting disease~\cite{liu2021monitoring},~\cite{low2020natural} and health-related information~\cite{shaw2022deciphering} from social media platforms. 

In our study for LDA topic modeling computation, we employed the use of Gensim~\cite{gensim}, an open-source Python library that specializes in unsupervised topic modeling. This library has been designed to extract semantic topics from a corpus of documents and can handle large text datasets efficiently. To distinguish between basic and advanced preprocessing techniques, we generated separate topic models for each setting. We calculated topics for different values of n (namely, 5, 10, 15, 20, 25, 30, 35, and 40) for both the basic and advanced preprocessing approaches. The topic models were visualized using word clouds and the pyLDAvis library in Python. The pyLDavis is an interactive web-based visualization tool that displays the topics deduced from a Latent Dirichlet Allocation (LDA) model, developed using a combination of the programming languages R and D3~\cite{sievert2014ldavis}. It offers a graphical user interface that makes it straightforward to understand the topic models and explore various visual representations of the data.

\section{EXPERIMENTS AND RESULTS}
In this section, we present the results generated from the analysis of our dataset using the NLP techniques outlined in the methodology which are, word clouds, topic modeling, and lexical analysis.
\subsection{Hyperparameters or Design Choices}
The coherence score indicates how closely related words occur within a topic, whereas the perplexity score indicates how effectively the model can predict previously unseen data using the trained model. A higher value of coherence score and a lower value of perplexity score is preferred, however, due to the data-driven nature of topic modeling, it is most important to consider the topics that have been discovered by the model and whether they are relevant for the task at hand. 

\begin{figure}[H]
  \centering
  \includegraphics[width=0.8\textwidth]{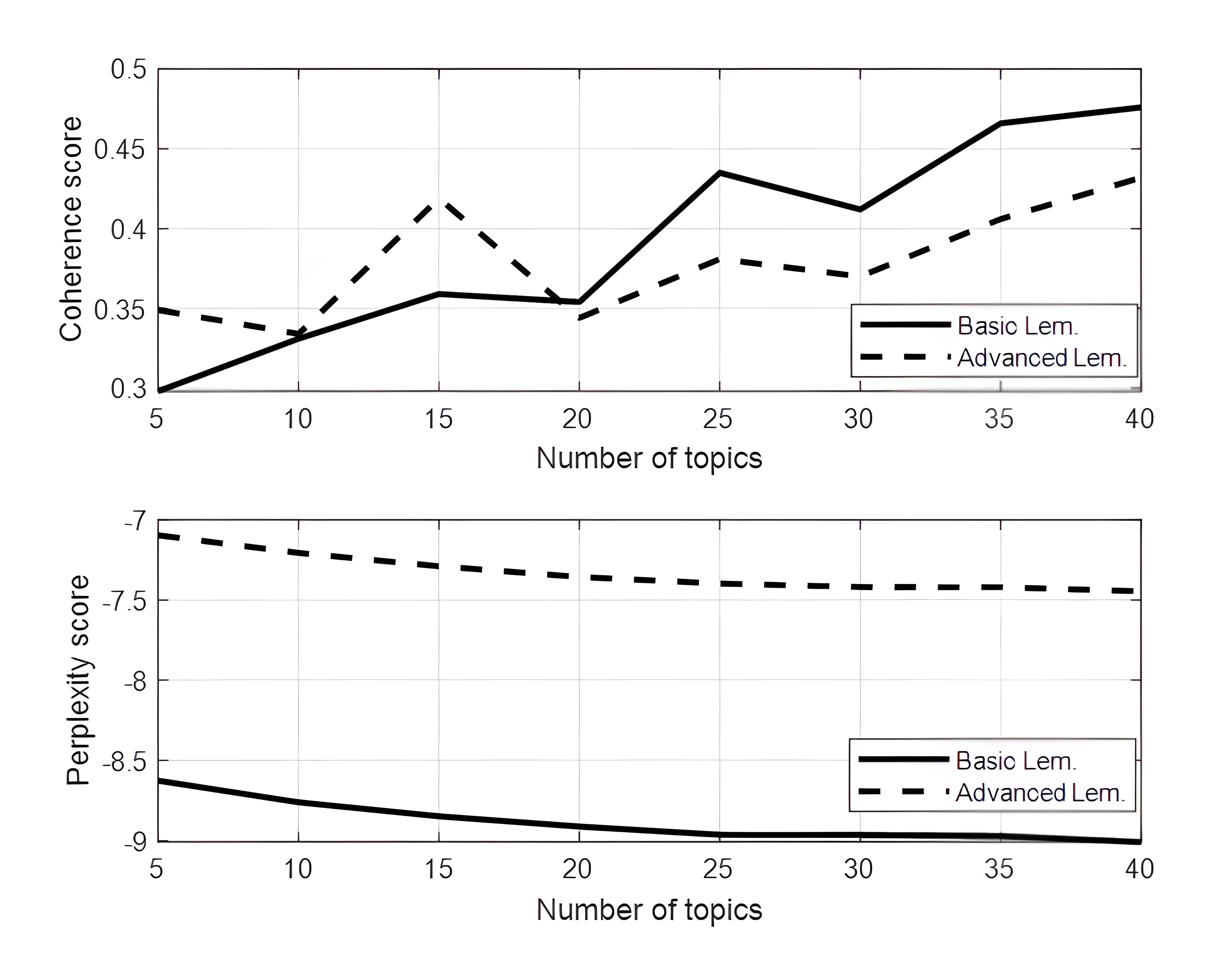}
  \caption{Perplexity and Coherence Score}
  \captionsetup{justification=centering} 
  \label{Perplexity}
\end{figure}

Table \ref{table4} presents the coherence and perplexity values for basic and advanced preprocessing with varying numbers of computed topics. From the table, it can be inferred that as the number of topics increases, the coherence value consistently rises for basic preprocessing, except for a slight dip when computed for 30 topics. However, for advanced preprocessing, the coherence value fluctuates for all topics. In contrast, the perplexity value consistently declines for both settings, indicating desirable results. The findings suggest that computing more topics increases the correlation between words within a specific topic and enhances the model's predictability when new data is introduced. Overall, the coherence and perplexity values for basic and advanced preprocessing, along with the number of computed topics, are presented in Table~\ref{table4} and Figure~\ref{Perplexity}, in tabular and graphical formats, respectively.	

\begin{table}[htp]
  \centering
  \caption{Coherence and Perplexity Scores of Preprocessing Methods}
  \begin{tabular}{cccccc}
    \toprule
    \textbf{\# Topics} & \multicolumn{2}{c}{\textbf{Basic Preprocessing}} & \multicolumn{2}{c}{\textbf{Advanced Preprocessing}} \\
    \cmidrule(lr){2-3} \cmidrule(lr){4-5}
     & \textbf{Coherence} & \textbf{Perplexity} & \textbf{Coherence} & \textbf{Perplexity} \\
    \midrule
    5   & 0.298 & -8.626 & 0.349 & -7.096 \\
    10  & 0.331 & -8.761 & 0.334 & -7.206 \\
    15  & 0.359 & -8.848 & 0.419 & -7.289 \\
    20  & 0.354 & -8.912 & 0.344 & -7.356 \\
    25  & 0.435 & -8.963 & 0.381 & -7.397 \\
    30  & 0.412 & -8.964 & 0.370 & -7.418 \\
    35  & 0.466 & -8.971 & 0.406 & -7.420 \\
    40  & 0.476 & -9.009 & 0.432 & -7.445 \\
    \bottomrule
  \end{tabular}
  \label{table4}
\end{table}

\renewcommand{\figurename}{Figure}
Given the wide variety of topics usually discussed in YouTube vlogs, with a smaller number of topics it may be difficult to adequately capture the entirety of the content. This in turn leads to a lower coherence score. However, if the number of topics is increased, there may not be sufficient data to fit this larger number of topics, thus diminishing the semantic value of the topics. Thus, it is critical to carefully evaluate the visual output of the topic models to determine the optimal number of topics that will capture the rich and diverse content present in YouTube vlogs.

\subsection{Lexical Analysis}
The lexicon of 200 words was categorized into 6 categories. From Table~\ref{tab:my-tablelexicon1} one can determine that words lying in the symptoms, generic terms, medical equipment’s and variants, and names of the virus category appeared at least in more than 50\% of the videos. The words in symptoms and generic terms categories appeared more prominent having 95.45\% and 96.46\% occurrences, respectively. It gives us the idea that most people were talking about symptoms and generic terms of COVID-19 in their videos. The words lying in the medicines and vaccines categories appeared with low percentages of 2.78\% and 4.42\%. We can assume that most of the videos we have gathered, date from a period when neither the public nor healthcare organizations were aware of what COVID-19 was, how harmful it could be to humans, which medications would be effective against it, and the vaccine had not yet been fully developed. Since there did not seem to be a remedy, the speakers in our data did not discuss much about the medications and vaccinations they were receiving. We also evaluated the percentage of transcripts in which the speakers were using words from the categories we have classified. From Table~\ref{table5} one can specify that there were only 0.76\% of videos in our dataset in which not a single word from our COVID-19 lexicon was used by the patients. In 37.17\% and 36.66\% of the transcripts, people were using terms of at least 3 and 4 categories, respectively. It shows us that our COVID lexicon provides great hindsight that the people in their videos were talking about coronavirus. 
\begin{table}[htp]
\centering
\caption{Percentage of transcripts containing words from the following categories}
\begin{tabular}{cc}
\toprule
\textbf{Categories} & \textbf{Percentage} \\
\midrule
Generic Terms & 96.46\% \\
Symptoms & 95.45\% \\
Medical Equipments & 65.11\% \\
Variants and Names & 52.59\% \\
Vaccines & 4.42\% \\
Medicines & 2.78\% \\
\bottomrule
\end{tabular}
\label{tab:my-tablelexicon1}
\end{table}
\begin{table}[htp]
\centering
\caption{Percentage of transcripts containing words from a different number of categories}
\begin{tabular}{cc}
\toprule
\textbf{Number of categories} & \textbf{Percentage} \\
\midrule
0 categories & 0.76\% \\
1 category & 2.91\% \\
2 categories & 18.96\% \\
3 categories & 37.17\% \\
4 categories & 36.66\% \\
\bottomrule
\end{tabular}
\label{table5}
\end{table}
\subsection{Word Clouds Results}
\begin{figure}[H]
\centering 
\includegraphics[width=0.8\textwidth]{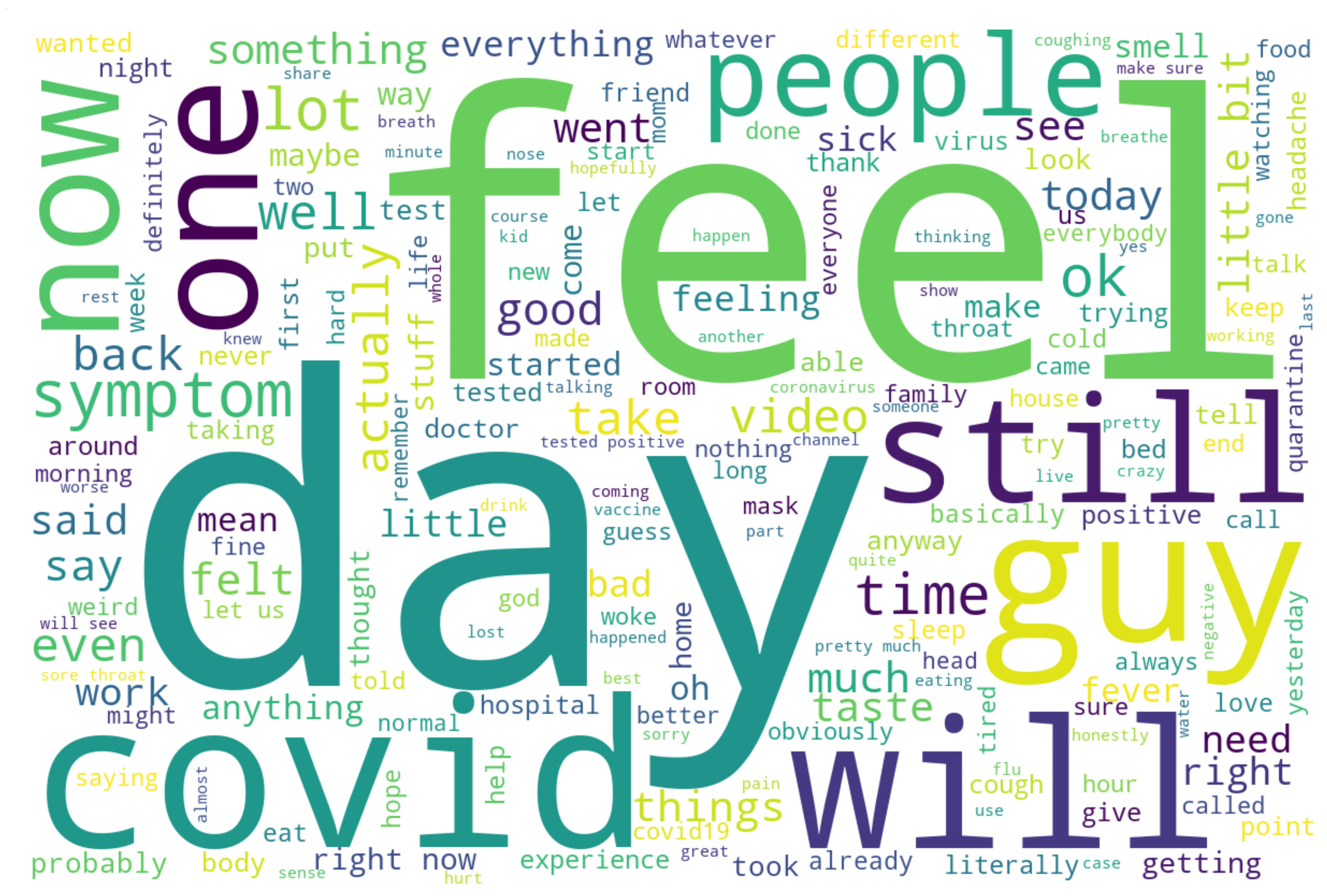} 
\captionsetup{justification=centering} 
\caption{Uni-gram Word Cloud including Stop Words}
\label{Unigramwithstop}
\end{figure}

\begin{figure}[H]
\centering 
\includegraphics[width=0.8\textwidth]{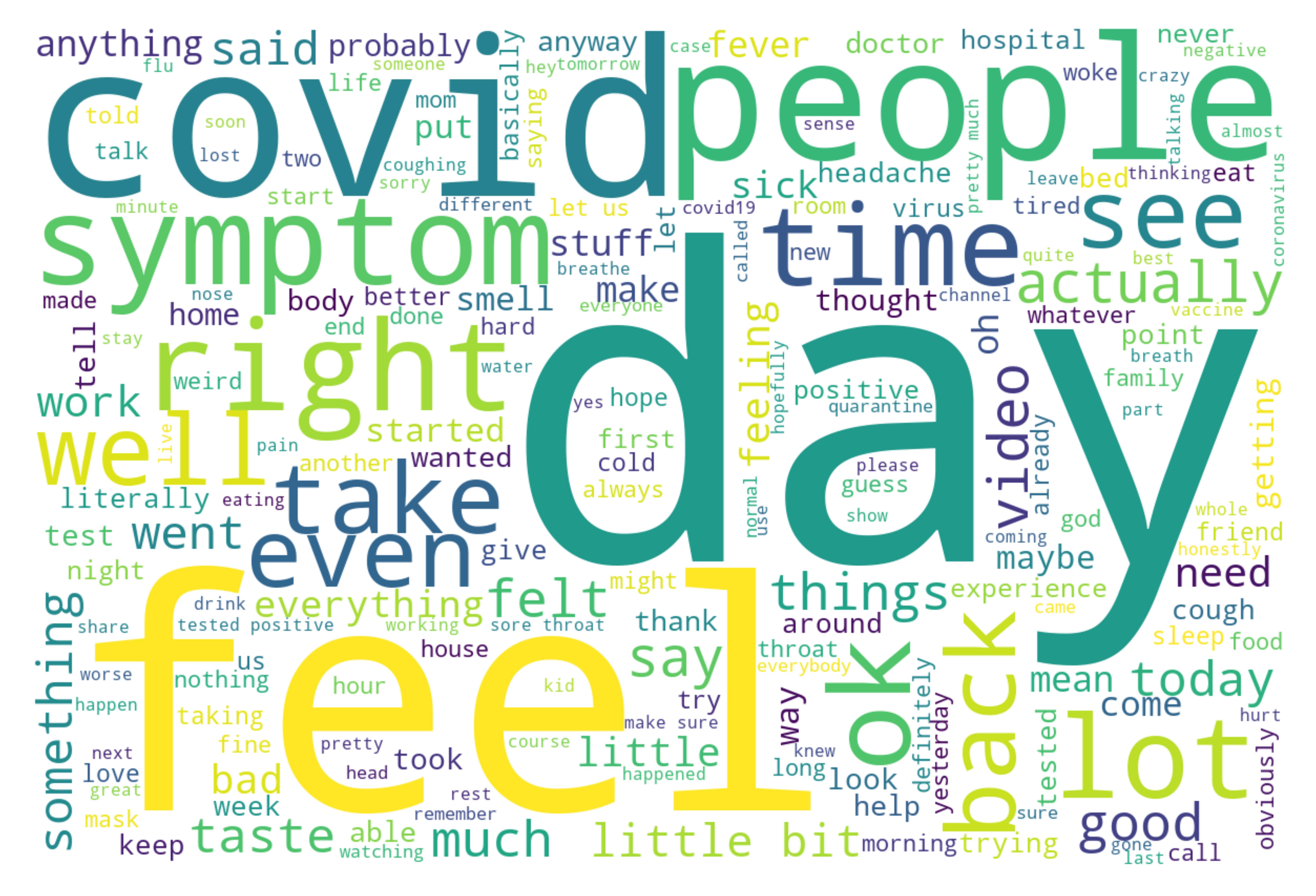} 
\captionsetup{justification=centering} 
\caption{Uni-gram Word Cloud after removing Stop Words}
\label{Unigramw}
\end{figure}

\begin{figure}[htp]
\centering 
\includegraphics[width=0.8\textwidth]{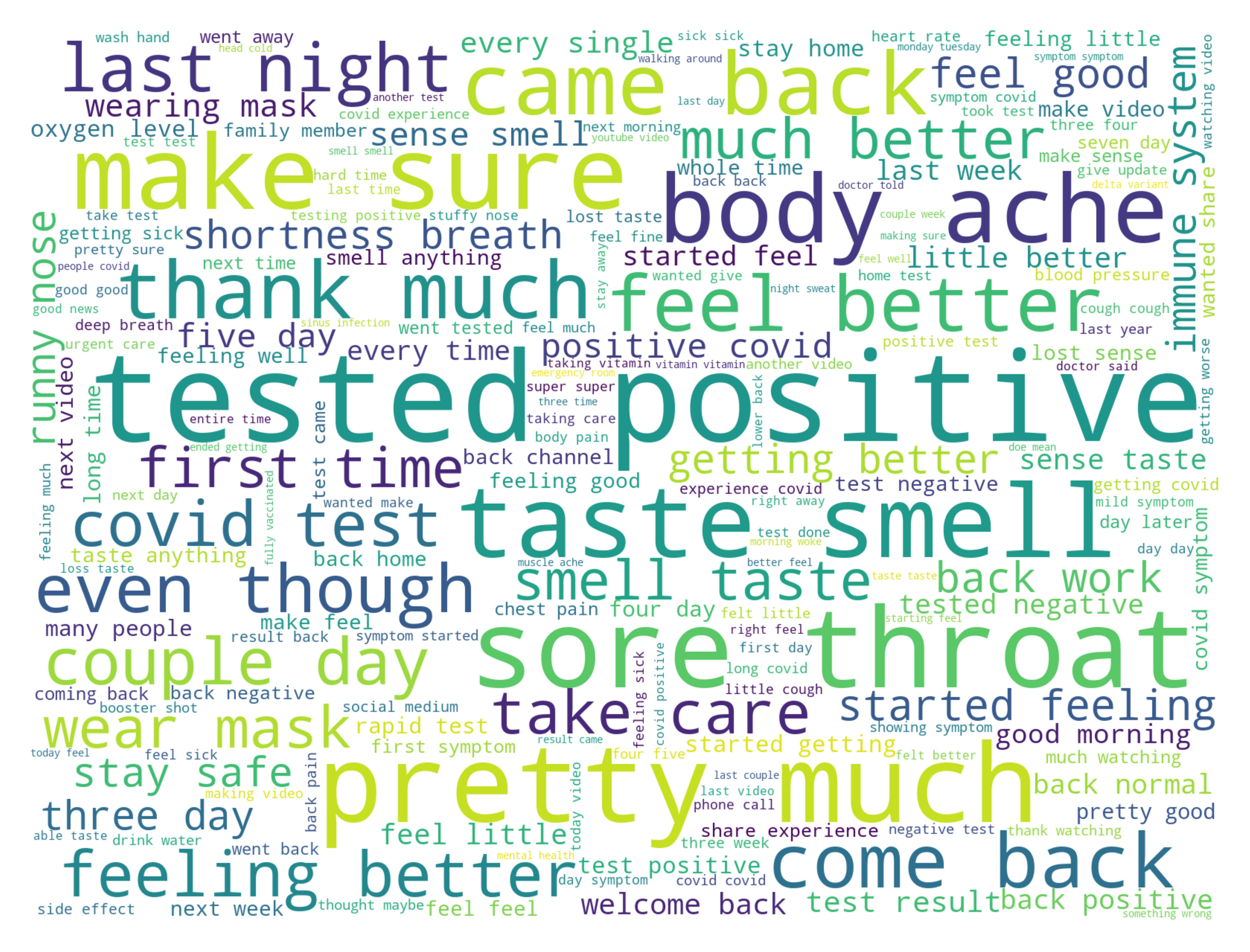} 
\captionsetup{justification=centering} 
\caption{Bi-gram Word Cloud}
\label{Bigram}
\end{figure}

To better understand the information contained in the COVID-19 related YouTube vlogs, we computed uni-gram, bi-gram, tri-gram, tetra-gram, and penta-gram word clouds. Figure~\ref{Unigramwithstop} illustrates the uni-gram word cloud in which single words are extracted from the COVID-19 vlogs data set. It includes relevant words such as "symptoms," "body," "COVID," "test," and "positive." In word clouds, the more used a word is in the data set the bigger its font would be, therefore in the uni-gram we can still see some words that do not supply us the necessary information but are large in size such as “day”, “thing”, “think”, “will”. Such words are called stop words and were removed to enhance the accuracy of the results and uni-gram word clouds were computed again as displayed in Figure~\ref{Unigramw}. 
\renewcommand{\figurename}{Figure}
\renewcommand{\figurename}{Figure}
Next, we generated the bi-gram word clouds as uni-gram does not provide substantial information. Figure~\ref{Bigram} displays a bi-gram word cloud that includes a combination of two words and provides a more thorough analysis of the topics people were discussing in their vlogs. One can see words such as “covid 19”, “test positive”, “sore throat”, “body ache”, “wear mask”, the frequency of stop words has decreased and more information can be interpreted from the bi-gram as compared to the uni-gram. 
\renewcommand{\figurename}{Figure}

\begin{figure}[H]
\centering 
\includegraphics[width=0.8\textwidth]{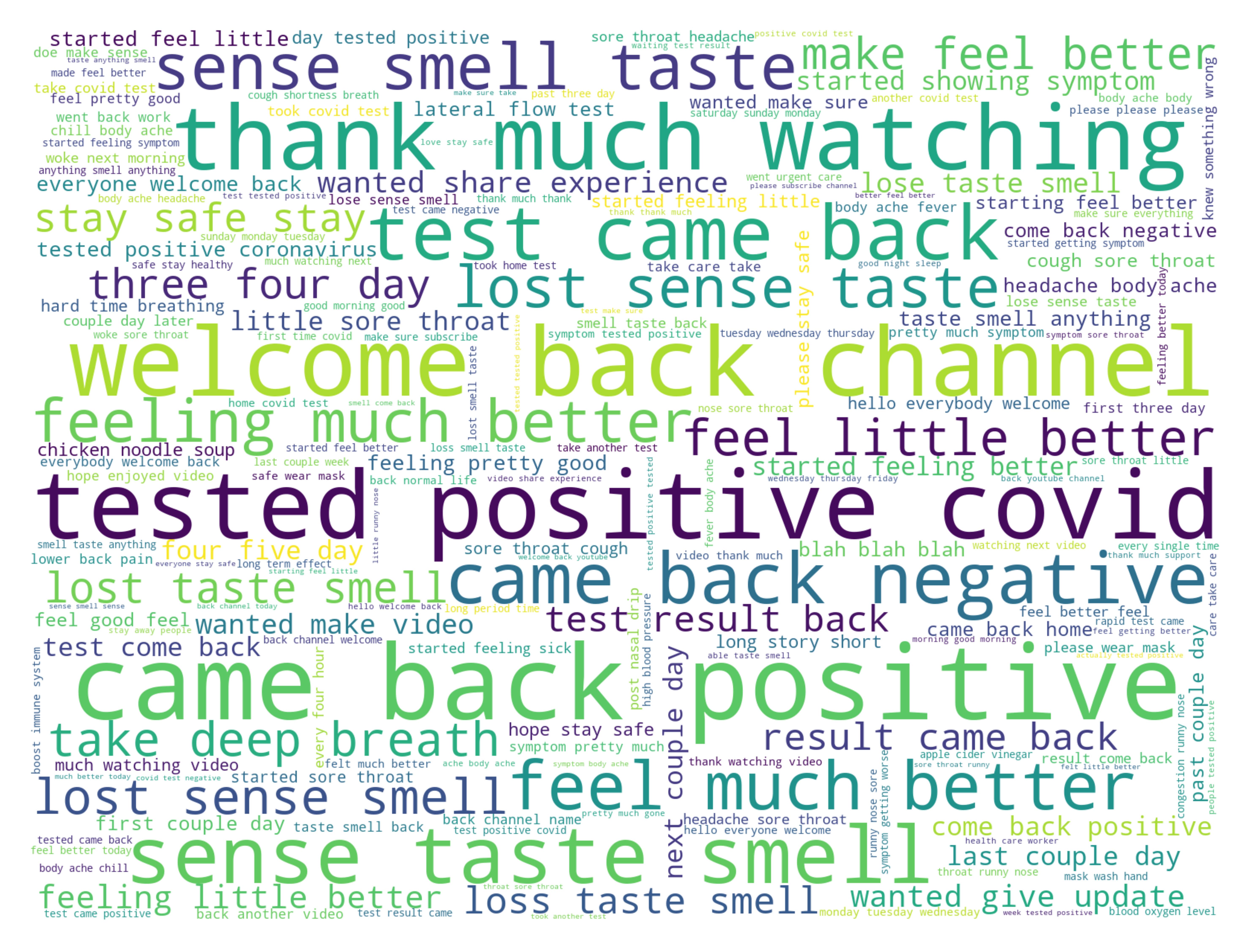} 
\captionsetup{justification=centering} 
\caption{Tri-gram Word Cloud}
\label{Trigram}
\end{figure}

\begin{figure}[H]
\centering 
\includegraphics[width=0.8\textwidth]{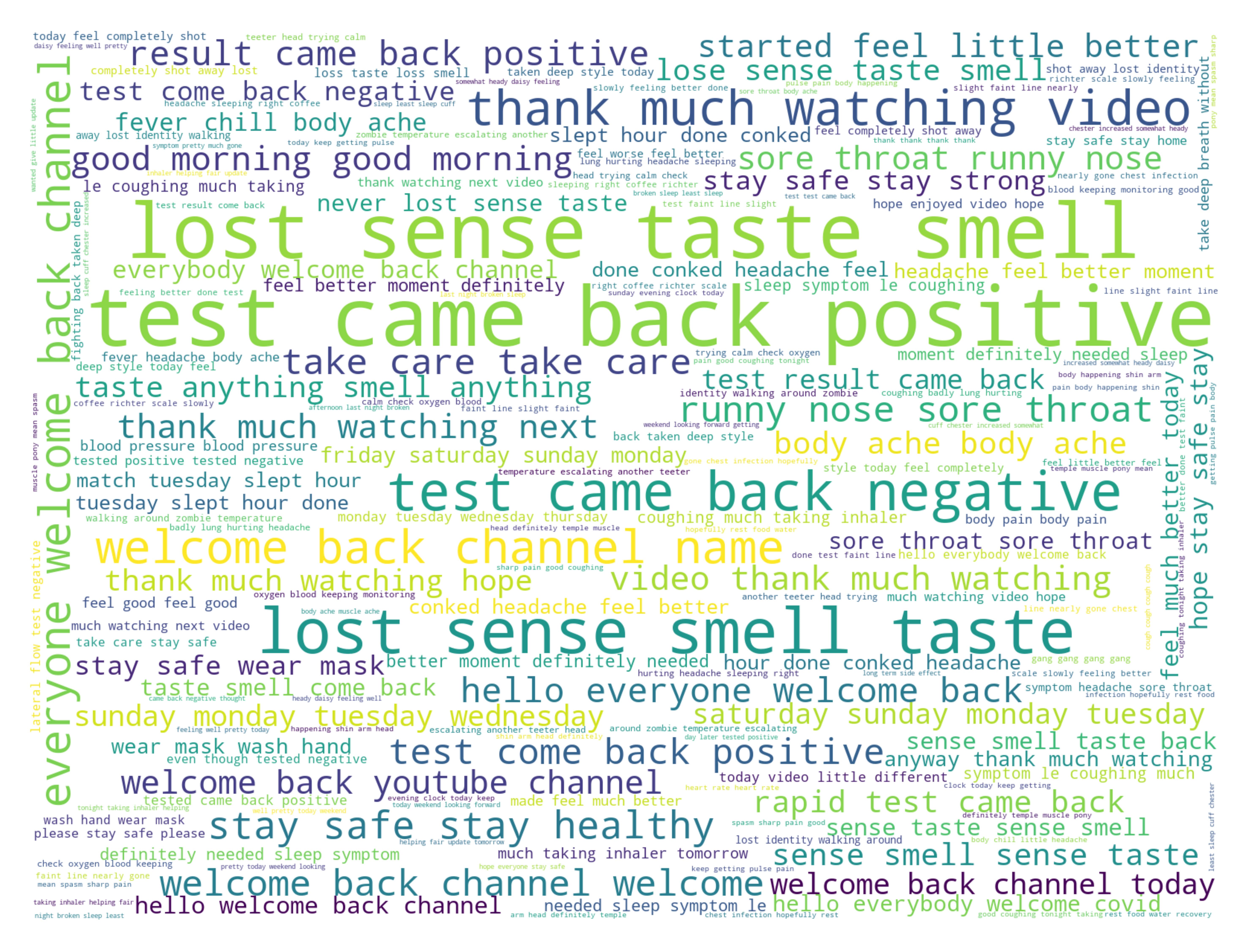} 
\captionsetup{justification=centering} 
\caption{Tetra-gram Word Cloud}
\label{Tetragram}
\end{figure}

\begin{figure}[H]
\centering 
\includegraphics[width=0.8\textwidth]{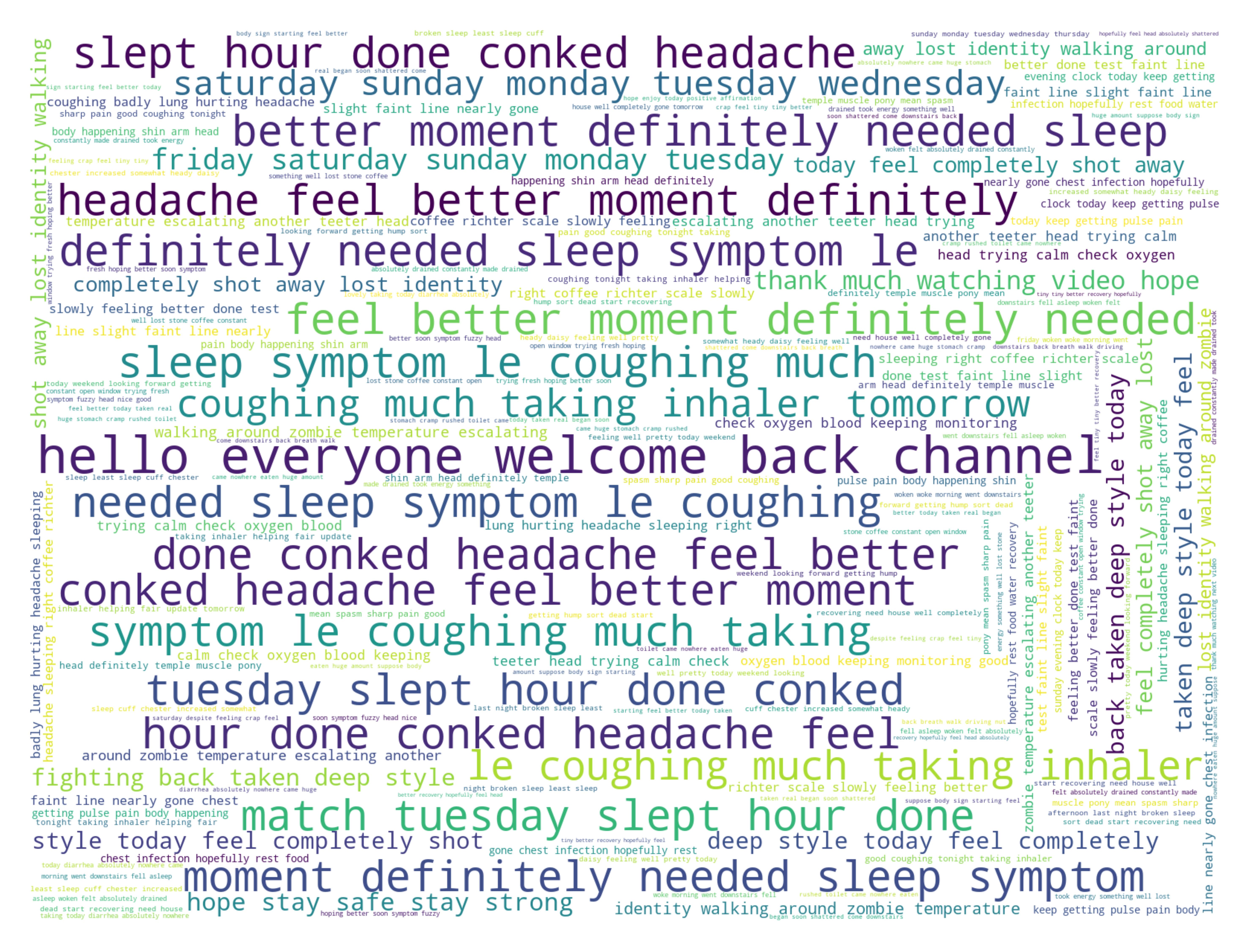} 
\captionsetup{justification=centering} 
\caption{Penta-gram Word Cloud}
\label{Pentagram}
\end{figure}

We further calculated the tri-gram which is a combination of three words and from Figure~\ref{Trigram} one can elucidate that there are phrases that are relevant to our study such as “test positive covid”, “feel much well”, “come back positive”, “come back negative”. We went ahead computing tetra-gram word clouds (a group of four words) and Figure~\ref{Tetragram} has terms that are pertinent to our research, such as “test positive covid 19”, “test come back positive”, and “loss sense taste smell” but the words are repeating as compared to those to the lower order word clouds. The penta-gram word cloud is presented in Figure~\ref{Pentagram}, however, it contains a lot of repetitive phrases and redundant information. 
\renewcommand{\figurename}{Figure}
\renewcommand{\figurename}{Figure}
\renewcommand{\figurename}{Figure}
We found that the information included in the word clouds starts to lose significance as we increase their order. While the uni-gram provides an initial overview, the bi-gram and tri-gram word clouds offer more detailed insights. However, beyond tri-gram, the word clouds tend to become repetitive and redundant, diminishing the significance of the information presented.

\subsection{Topic Modeling Results}
\begin{figure}[htp]
  \centering
  \includegraphics[width=1.0\textwidth]{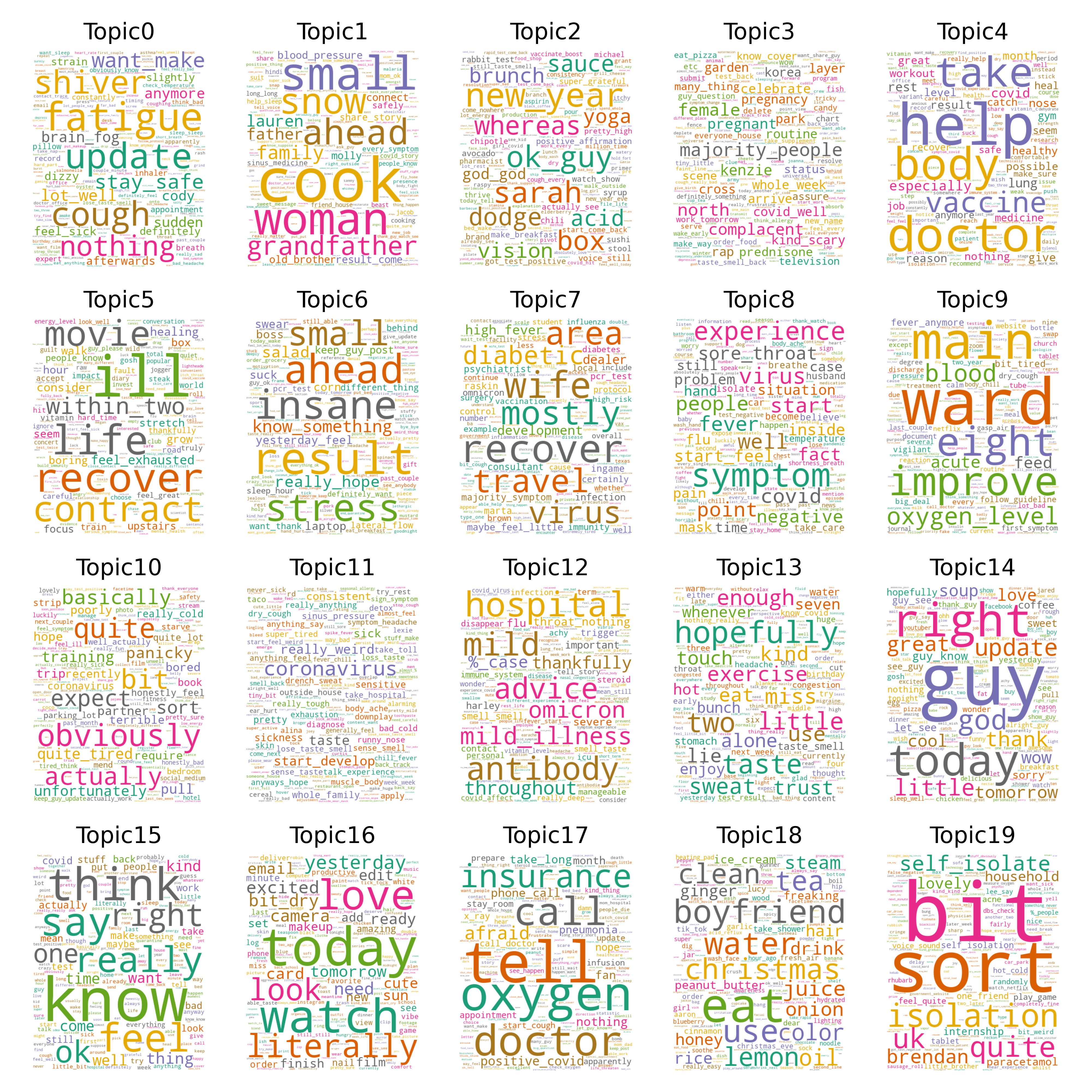}
  \caption{Basic Preprocessing LDA Topic Modeling (n=20)}
  \captionsetup{justification=centering} 
  \label{Basic20}
\end{figure}

For topic modeling, a well-known technique called Latent Dirichlet Allocation (LDA) was employed. We computed topics ranging from 5 to 40 with an increment of 5 using both basic and advanced preprocessing methods. We analyzed all the computations and interpreted that using a lower number of topics does not adequately capture the entirety of the textual data and increasing topics beyond a certain point causes a loss of information as there will not be enough data to support that many topics. As a result, we found that using a moderate number of topics provided valuable insights into the textual data which are illustrated in Figure~\ref{Basic20}. 

We analyzed that the results from basic preprocessing topics (n=20) provide a comprehensive understanding of the topics related to the COVID-19 pandemic, as shown in Figure 16. Topic 0 revealed that people discussed general symptoms associated with their experiences with COVID-19, such as “shiver”, “cough”, “fatigue”, and “dizzy”. On the other hand, topic 8 is specifically related to the coronavirus, with terms like “virus”, “symptom”, “negative”, and “covid” being prevalent. Furthermore, topics 9 and 17 were associated with hospitals and medical terms such as “ward”, “blood”, “oxygen level”, “doctor”, “appointment”, “call”, and “insurance”, indicating that the public was concerned about their health during the pandemic and was coordinating with medical professionals and insurance companies. In addition, topic 18 revealed that people were also relying on home remedies for relief from their symptoms due to the misinformation surrounding the pandemic, with terms like “onion”, “lemon”, “honey”, “oil”, “tea”, “ginger”, and “steam” being commonly discussed. 
\renewcommand{\figurename}{Figure}

\begin{figure}[htp]
  \centering
  \includegraphics[width=1.0\textwidth]{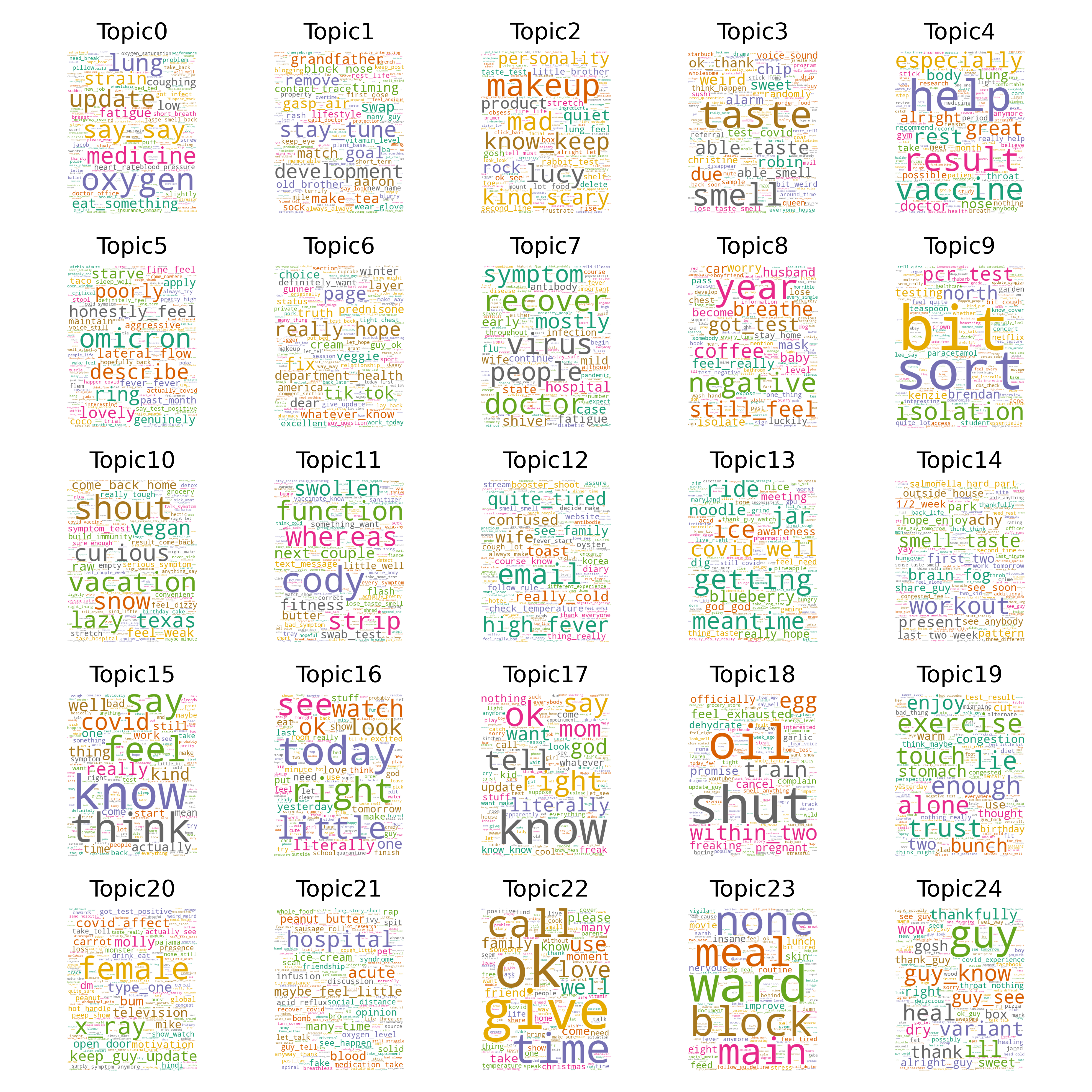}
  \caption{Basic Preprocessing LDA Topic Modeling (n=25)}
  \captionsetup{justification=centering} 
  \label{Basic25}
\end{figure}

To assess the potential benefits of increasing the number of topics analyzed, we expanded our calculations to include 25 topics. The results of this analysis are shown in Figure~\ref{Basic25}. From topic 7 one can see generic terms related to COVID-19 such as “symptom”, “recover”, “virus”, “people”, and “doctor”. Topic 9 is associated with getting tested and being in isolation as terms like “pcr\_test”, “isolation”, “testing”, and “netflix” manifest our idea. Moreover, topic 13 seems connected with what people were eating during their period of isolation, as in this topic one can see terms such as “blueberry”, “ice”, “noodle”, and “jar”. The interpretation from the Figure~\ref{Basic25} suggests that although the model with 25 topics offers valuable information, their quality is not as good as the earlier set with 20 topics. This is mainly because many words in most of the topics do not seem to have any significant connection with each other, and repetition of words is also seen.
\renewcommand{\figurename}{Figure}

\begin{figure}[htp]
  \centering
  \includegraphics[width=1.0\textwidth]{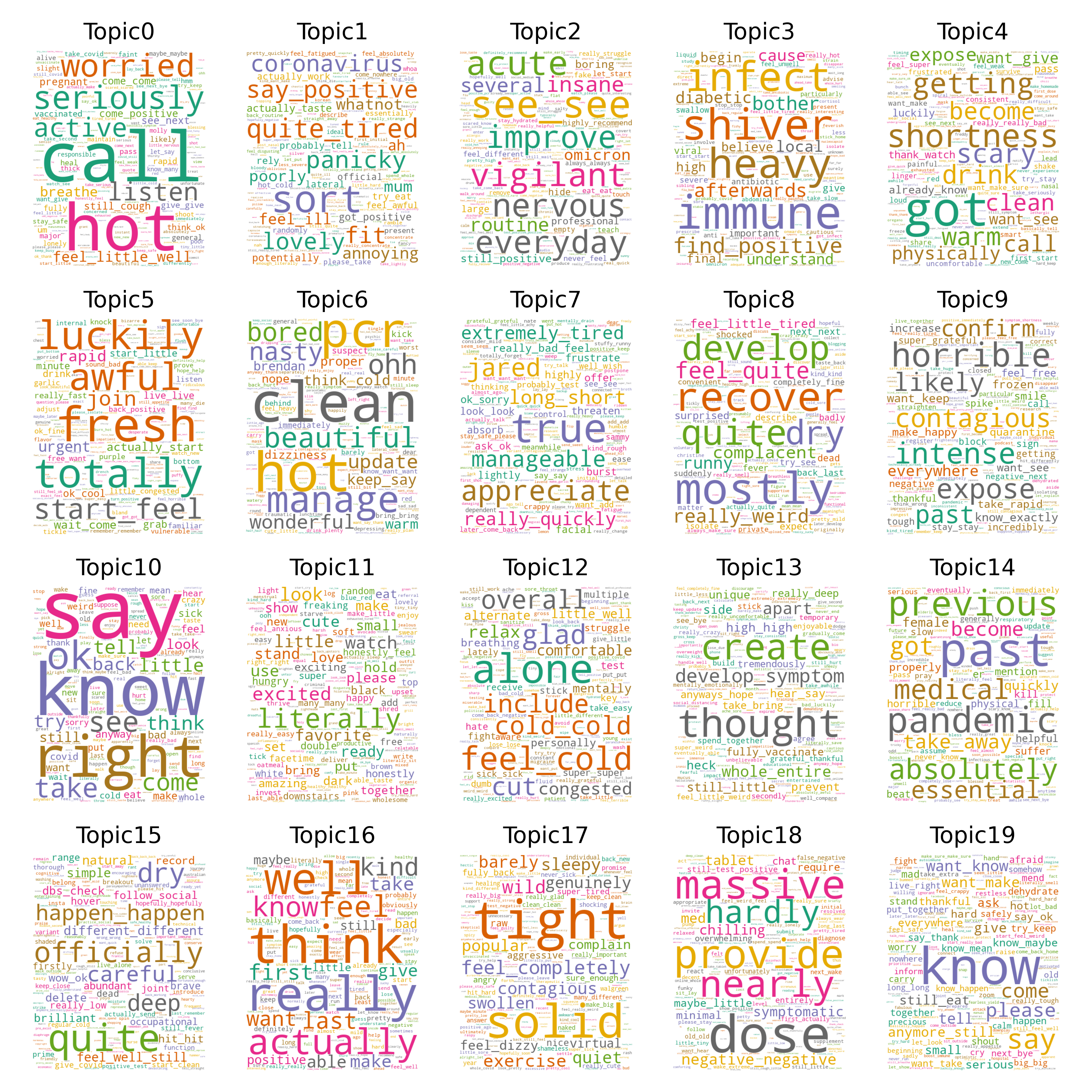}
  \caption{Advanced Preprocessing LDA Topic Modeling (n=20)}
  \captionsetup{justification=centering} 
  \label{Advanced20}
\end{figure}

\begin{figure}[H]
  \centering
  \includegraphics[width=0.8\textwidth]{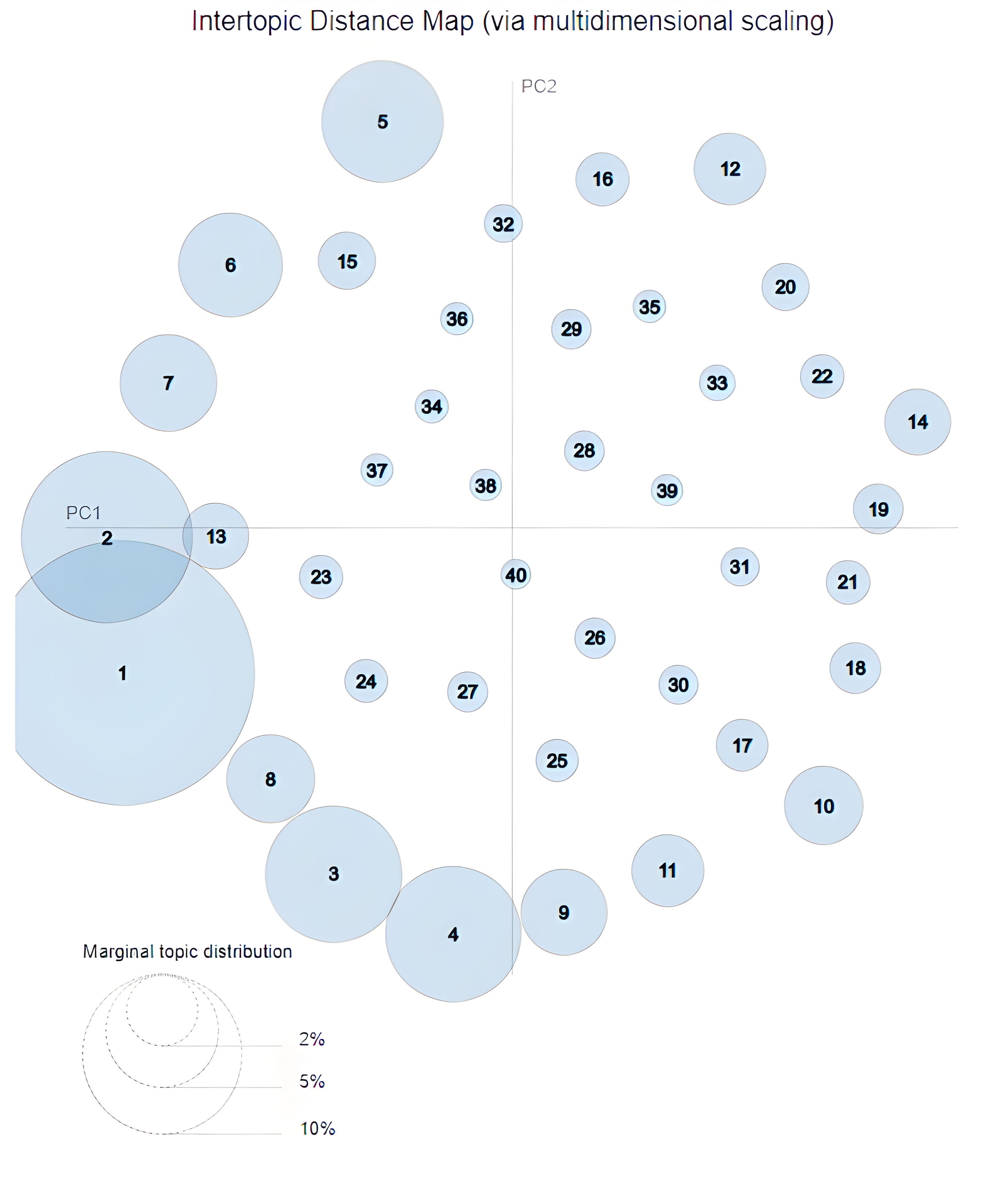}
  \caption{Intertopic Distance Map (Advanced Preprocessing)}
  \captionsetup{justification=centering} 
  \label{Intertopic}
\end{figure}

\begin{figure}[htp]
  \centering
  \includegraphics[width=0.8\textwidth]{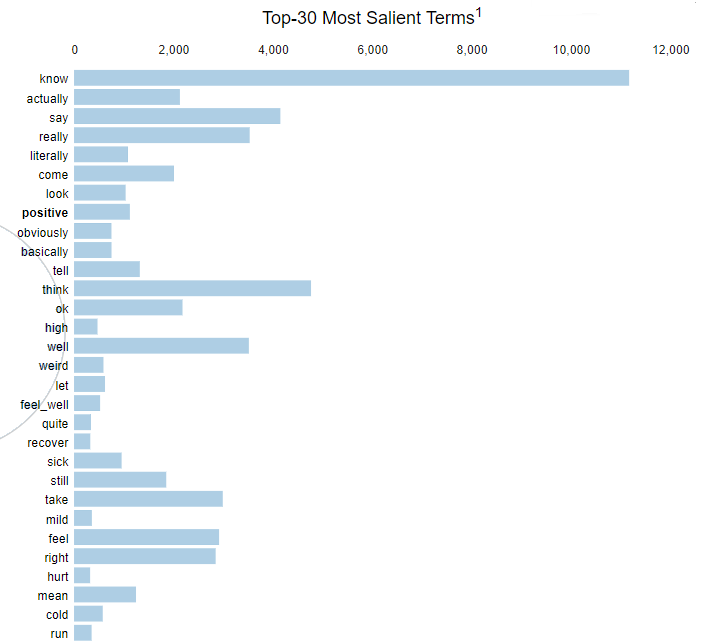}
  \caption{Top 30 Salient Terms in Advanced Preprocessing}
  \captionsetup{justification=centering} 
  \label{Top30}
\end{figure}

The results extracted from advanced preprocessing with 20 topics are illustrated in Figure~\ref{Advanced20}. If we increase the number of topics in the advanced preprocessing framework, we tend to lose the desired information, while a lesser number also does not provide us with valuable insights. Therefore, we chose 20 topics as it allows us to get a better understanding of people’s experiences related to the virus. The results reveal that topics 2 and 3 contain general terms related to the pandemic that people in their vlogs such as “coronavirus”, “say\_positive”, “quite\_tired”, “feel\_ill”, “infect”, “immune”, “find\_positive”, and “shiver”. Moreover, topic 9 presents terms related to the spreading and extent of the virus like “contagious”, “horrible”, “intense”, “expose”, “likely”, and “everywhere”. The public's concern about the virus spread is evident from this result. However, we believe basic preprocessing yields better results as compared to advanced preprocessing.
\renewcommand{\figurename}{Figure}
\newpage
The intertopic distance map is illustrated in Figure~\ref{Intertopic} and it was computed using the pyLDAvis library. This map provides a comprehensive analysis of the topics under consideration. These maps were computed for both basic and advanced preprocessing settings. However, the advanced preprocessing setting was examined to provide more distinct topics. This is attributed to the fact that it only included significant words such as verbs, adjectives, adverbs, and interjections while eliminating all other types of words that were considered irrelevant. The largest topic on the map is Topic 1, which consists of common terms related to the study. While topics 13, 18, 15, and 3 overlap, indicating a correlation between their terms, the remaining topics are distinct from one another, with their respective topic numbers being separated by distance. Overall, the intertopic distance map is a useful method for visualizing and analyzing the relationships between topics and their associated terms.
\renewcommand{\figurename}{Figure}
Figure~\ref{Top30} displays the 30 most significant terms in all 40 topics. It reveals that terms such as “positive”, “recover”, “sick”, “mild”, “cold”, and “feel” are relevant to our research and frequently appear in the topics. This supports our idea that individuals in their vlogs were discussing their experiences of testing positive for the virus, recovering from it, and its impact on their health.
\renewcommand{\figurename}{Figure}

\renewcommand{\figurename}{Figure}
\section{CONCLUSION}
\subsection{Discussion}
These days, people often use social media platforms for sharing their health experiences and gaining medical information. In our research, we proposed a natural language processing based method for analyzing the content related to COVID-19 videos available on YouTube. The topic modeling, lexicon analysis, and word clouds results were generated automatically and provided a comprehensive overview of what people were talking about and experiencing when this deadly virus affected them. As one can see in our LDA topic modeling results, the individuals in the vlogs were discussing their COVID-19 information and the related words have been grouped into specific topics. People shared information about the symptoms they experienced and discussed the precautions and home remedies they used to reduce the impact of the virus. The topic modeling results also highlight the public's worries about scheduling appointments with doctors and their health insurance, providing crucial information about the public's response to COVID-19. Our lexicon results signify the accuracy of our methods, as only 0.76\% of the videos we downloaded did not contain a term from the 200 words COVID-19 lexicon we created. The word cloud results provide valuable insight into the most used terms related to the virus which individuals shared in the YouTube vlogs, as illustrated in the n-grams presented above.

We believe that our results can be used as a motivation to create and test an automatic system where people would share their symptoms related to COVID-19, and the system using an automatic speech recognition tool would convert their speech into a textual format and find whether they are suffering from COVID-19. This study highlights the potential of NLP as a tool for public health screening and disease surveillance on YouTube. Utilizing NLP would also relieve us from the time-consuming way of manually keeping health records of individuals, usually done by health organizations. 

The findings of our research provide a conclusive response to our first research question demonstrating that NLP can be a useful tool in identifying COVID-19 symptoms from YouTube vlogs. The results from our NLP techniques including topic modeling, and word clouds revealed that people in the vlogs were discussing a wide range of COVID-19 symptoms such as fever, flu, cough, sore throat, etc. Additionally, our lexical analysis results showed that COVID-19 symptoms were discussed in 95.45\% of transcript files generated from YouTube vlogs. Moreover, our results address our second question that by using NLP through YouTube vlogs, one can identify the symptoms, general terms, medicines, medical equipment, and vaccines that would be vital for public health professionals and policymakers in controlling and mitigating the spread and effect of viral diseases. 
\subsection{Key Takeaways and Insights}
When COVID-19 officially became a pandemic, there were a lot of rumors and misinformation circling around the extent of this virus and what effects it could have on the public. One way to understand viral infections is through patient history, but hospitals manually record patient symptoms, which can be time-consuming. Our study reveals that an automatic system could be created to check the symptoms of COVID-19 by automatically analyzing the transcripts using Natural Language Processing techniques. This information could be used for the social benefit and policymaking of the world to be prepared if it hinders people in the future. Our findings reveal that the transcripts of people’s personal experiences of COVID-19 provide a reliable source of information on the symptoms, general terms, medicines, and precautions associated with this pandemic. We believe that NLP is effective in understanding human communication in an automatic manner and could be a vital tool for public health screening of COVID-19 and other infectious diseases.

\subsection{Limitations}
Since anyone can upload videos on YouTube, it is possible there may have been people who were making up their symptoms and spreading false information. Also, for a clearer and more comprehensive view of COVID-19, we could have created a larger dataset. In addition, there is a large volume of videos on YouTube, therefore it is possible we overlooked some of them that would have been useful for our research. Some of the videos we downloaded were deleted by the users which created a loss of useful information in our study as then we were not certain that the videos were from a channel that contains vlogs.

\subsection{Implications for Practitioners and Policymakers}
The findings of our study could have practical applications for practitioners and policymakers who are tasked with keeping track of and responding to trends in public discourse around infectious diseases. Our dataset and methods could prove to be vital for policymakers and practitioners who can use them for public health screening of COVID-19 and to combat the spread of the virus. Through the suggested methods public health practitioners could figure out the most common symptoms met by the patients and create targeted public health campaigns to increase awareness about these symptoms and motivate individuals to seek medical aid if they experience them. In addition, by using our dataset they could set up strategies to address the misinformation which has been spread among the public about this virus and make sure that accurate information is readily available to them. Overall, our study underscores the importance of NLP on YouTube vlogs for public health screening of COVID-19 and infectious diseases and how it could be employed in minimizing the effects of such diseases. 

\subsection{Directions for Future Work}
For speech-to-text conversion, currently, we have only used Microsoft Streams as an ASR tool. Future research could be aimed at examining the usage of other ASR tools such as IBM Watson, Microsoft Azure, and Google Speech to Text. By comparing the accuracy and quality of textual data generated by these tools, the most effective ASR tool could be identified. In addition, our dataset could be enlarged by including flu videos, to compare the similarities and differences between COVID-19 and flu datasets. The research could be conducted in assessing whether certain symptoms or behaviors are mentioned more frequently in one dataset compared to another. The discovery of such differences may offer valuable information about distinctive features of COVID-19 discussion, which could guide future research on the subject. In addition, we would share our lexicon and transcripts files after removing identifiable information for future research on this topic. 

\section*{Declarations}
\subsection*{Ethical Approval}
Ethical approval was not required for the use of YouTube originated data, as it is publicly available.
\subsection*{Competing Interests}
We declare that we have no known competing financial interests or personal relationships that are directly or indirectly related to this paper. 
\subsection*{Author Contributions}
Ahrar Bin Aslam collected data, conceived and designed the experiments, conducted the experiments, analyzed the data, created figures and/or tables, contributed to writing and reviewing drafts of the paper, and approved the final version of the draft.
Zafi Sherhan Syed conceived and designed the experiments, conducted the experiments, analyzed the data, created figures and/or tables, contributed to writing and reviewing drafts of the paper, and approved the final version of the draft.
Muhammad Shehram Shah Syed conceived and designed the experiments, conducted the experiments, analyzed the data, created figures and/or tables, contributed to writing and reviewing drafts of the paper, and approved the final version of the draft.
Muhammad Faiz Khan collected data, conducted the experiments, created figures and/or tables, contributed to writing and reviewing drafts of the paper, and approved the final version of the draft.
Asghar Baloch collected data, conducted the experiments, created figures and/or tables, contributed to writing and reviewing drafts of the paper, and approved the final version of the draft.
\subsection*{Funding}
Not Applicable.
\subsection*{Availability of Data and Materials}
To ensure privacy of individuals, we shall not distribute audio or video recordings. If required, we shall release anonymised transcripts for academic research. 
\bibliography{sn-article.bib} 

\end{document}